\documentclass{article}

\usepackage[preprint]{cpal_2025}

\usepackage{url}
\usepackage[utf8]{inputenc} % allow utf-8 input
\usepackage[T1]{fontenc}    % use 8-bit T1 fonts
\usepackage{hyperref}       % hyperlinks
\usepackage{url}            % simple URL typesetting
\usepackage{booktabs}       % professional-quality tables
\usepackage{amsfonts}       % blackboard math symbols
\usepackage{nicefrac}       % compact symbols for 1/2, etc.
\usepackage{microtype}      % microtypography
\usepackage{xcolor}         % colors
\usepackage{amsmath}
\usepackage[capitalize,noabbrev]{cleveref}
\usepackage{graphicx}
\usepackage{subfigure}
\usepackage{algorithm}
\usepackage{algorithmic}

\title{Closure Discovery for Coarse-Grained Partial Differential Equations Using Grid-based Reinforcement Learning}

\author{%
Jan-Philipp von Bassewitz$^{1,2}$ \quad Sebastian Kaltenbach$^{1,2}$ \quad Petros Koumoutsakos$^2$ \thanks{Corresponding Author: \texttt{petros@seas.harvard.edu}}\\
$^1$ETH Zurich \quad $^2$Harvard SEAS}
\begin{document}

\maketitle

\begin{abstract}
Reliable predictions of critical phenomena, such as weather, wildfires and  epidemics often rely  on models described by Partial Differential Equations (PDEs). However, simulations that capture the full range of spatio-temporal scales described by such  PDEs are often prohibitively expensive. Consequently, coarse-grained simulations  are usually deployed that adopt various  heuristics and empirical closure terms to account for the missing information. We propose a novel and systematic approach for identifying closures in under-resolved PDEs using grid-based Reinforcement Learning. This formulation incorporates inductive bias and exploits locality by deploying a central policy represented efficiently by a Fully Convolutional Network (FCN). 
We demonstrate the capabilities and limitations of our framework through numerical solutions of the advection equation and the Burgers' equation. Our results show accurate predictions for in- and out-of-distribution test cases as well as a significant speedup compared to resolving all scales.
\end{abstract}

\section{Introduction}
Simulations of critical phenomena such as climate,  ocean dynamics and epidemics, have become essential for decision-making, and their veracity, reliability, and energy demands have great impact on our society. Many of these simulations are based on models described by PDEs expressing system dynamics that span multiple spatio-temporal scales. Examples include  turbulence~\cite{wilcox1988multiscale}, neuroscience~\cite{dura2019netpyne}, climate~\cite{climatenas} and  ocean dynamics~\cite{mahadevan2016impact}.

Today, we benefit from decades of remarkable efforts in the development of  numerical methods, algorithms, software, and hardware and witness simulation frontiers that were unimaginable even a few years ago \cite{hey2009fourth}. 
Large-scale simulations that predict the system's dynamics may use trillions of computational elements \cite{rossinelli2013a} to resolve all spatio-temporal scales, but these often address only idealized systems and their computational cost prevents experimentation and uncertainty quantification.

By contrast, reduced order and coarse-grained models are fast, but limited by the linearization of complex system dynamics while  their associated closures, which model the effect of unresolved (not simulated) dynamics on the quantities of interest, are in general based on heuristics and, as a result, domain specific \cite{peng2021multiscale}. A closure discovery framework that is independent of the system of interest and can be applied to various domains and tasks is thus highly desirable \cite{sanderse2024scientific}.
%A simulation of the aforementioned systems is challenging as the smallest spatial and temporal scales relevant for the systems need to be resolved. Additionally, most multi-scale systems are also multi-physics and the different processes at each scales involve different physical descriptions and models \cite{peng2021multiscale}. For some systems, the physics involved at small scales is even partially unknown and thus a direct simulation is not possible.\\

To address this challenge we propose \textit{Closure-RL}, a framework to complement coarse-grained simulations with closures that are discovered by a grid-based Reinforcement Learning (RL) framework. Closure-RL employs a central policy that is based on a FCN and uses locality as an inductive bias. It is able to correct the error of the numerical discretization and improves the overall accuracy of the coarse-grained simulation. 

Our approach is inspired by pixelRL \cite{cv_pixel_rl}, a recent work in Reinforcement Learning (RL) for image reconstruction, which minimizes local reconstruction errors. PixelRL employs a \textit{per-pixel} reward and value network and can therefore be interpreted as a cooperative Multi-Agent Reinforcement Learning (MARL) framework with one agent per pixel. We extend this framework to closure discovery by treating the latter as a reconstruction problem. The numerical scheme introduces corruptions, that the agents are learning to reverse. In contrast to actions based on a set of filters as in  pixelRL, we employ a continuous action space that is independent of the PDE to be solved. Our agents learn and act locally on the grid, in a manner that is reminiscent of the numerical discretizations of PDEs based on Taylor series approximations. Hereby, each agent only gets information from its spatial neighborhood. 

Although Closure-RL has many characteristics of a cooperative MARL approach, the training costs are similar to a single-agent RL formulation as our proposed formulation employs a central policy and does not require complex interactions between the agents. This allows us to use a large number of agents and no fine-tuning regarding the placement of the agents is required. After training, the framework is capable of accurate predictions in both in- and out-of-distribution test cases. We remark  that the actions taken by the agents are highly correlated with the numerical errors and can be viewed as an implicit correction to the effective convolution performed via  the coarse-grained discretization of the  PDEs \cite{bergdorf2005multilevel}.
We note that we have chosen a RL-based approach to not require access to a differentiable solver or potentially difficult gradient computations via the adjoint method \cite{sanderse2024scientific}. Our approach is non-intrusive and can be seamlessly integrated with any numerical discretization scheme. %\textit{Our contribution can therfore be summarized as an end-to-end RL closure discovery framework that is independent of the PDE, does not require access to the solver (to obtain gradients) and, in contrast to other RL approaches, also works offline with pre-computed high-resolution trajectories.}

The main contribution of this work is a grid-based RL algorithm for the discovery of closures for coarse-grained PDEs. The algorithm provides an automated process for the correction of the errors of the associated numerical discretization. We show that the proposed method is able to capture scales beyond the training regime and provides a potent method for solving PDEs with high accuracy and limited resolution.

\section{Related Work}
\textbf{Machine Learning and Partial Differential Equations:}
In recent years, there has been significant interest in learning the solution of PDEs using Neural Networks. Techniques such as 
PINNs \cite{raissi2019physics,karniadakis2021physics}, DeepONet \cite{lu2021learning}, the Fourier Neural Operator \cite{li2020fourier}, NOMAD \cite{seidman2022nomad}, Clifford Neural Layers \cite{brandstetter2022clifford} and an invertible formulation \cite{kaltenbach2023semi} have shown promising results for both forward and inverse problems. However, there are concerns about their accuracy and related computational cost, especially for low-dimensional problems \cite{karnakov2022optimizing}. These methods aim to substitute numerical discretizations with neural nets, in contrast to our RL framework, which aims to complement them. Moreover, their loss function is required to be differentiable, which is not necessary for the stochastic formulation of the RL reward.\\
\textbf{Reinforcement Learning:}
The present approach is designed to solve various PDEs using a central policy and it is related to similar work for image reconstruction \cite{cv_pixel_rl}. Its efficient grid-based formulation is in sharp contrast to multi-agent learning formulations that train agents on decoupled subproblems or learn their interactions \cite{yang2020overview,freed2021learning,wen2022multi,albrecht2023multi,sutton2023reward}. The single focus on the local discretization error allows for a general method that is not required to be fine-tuned for the actual application by selecting appropriate global data (such as the energy spectrum) \cite{novati2021automating} or domain-specific agent placement \cite{bae2022scientific}. \\
\textbf{Closure Modeling:}
The development of machine learning methods for discovering closures for under-resolved  PDEs has gained attention in recent years. Current approaches are mostly tailored to specific applications such as turbulence modeling \cite{ling2016reynolds, durbin2018some,novati2021automating,bae2022scientific} and use data such as energy spectra and drag coefficients of the flow in order to train the RL policies. In \cite{lippe2023pde}, a more general framework based on diffusion models is used to improve the solutions of Neural Operators for temporal problems using a multi-step refinement process. Their training is based on supervised learning in contrast to the present RL approach which additionally complements existing numerical methods instead of neural network based surrogates \cite{li2020fourier,gupta2022towards}.\\
\textbf{Inductive Bias:}  
The incorporation of  prior knowledge regarding the physical processes described by the PDEs, into machine learning algorithms is critical for their training in the Small Data regime and for increasing the accuracy during extrapolative predictions \cite{goyal2022inductive}. One way to achieve this is by shaping appropriately the loss function \cite{karniadakis2021physics, kaltenbach2020incorporating,yin2021augmenting, wang2021learning, wang2022respecting}, or by incorporating parameterized mappings that are based on known constraints \cite{greydanus2019hamiltonian,kaltenbach2021physics, cranmer2020lagrangian}. Our RL framework incorporates locality and is thus consistent with numerical discretizations that rely on local Taylor series based approximations. The incorporation of  inductive bias in RL has also been attempted by focusing on the beginning of the training phase \cite{uchendu2023jump,walke2023don} in order to shorten the exploration phase.

\section{Methodology}
We consider a time-dependent, parametric, non-linear PDE defined on a regular domain $\Omega$. The solution of the PDE depends on its initial conditions (ICs) as well as its PDE-parameters (PDEP) $ C \in \mathcal{C}$. The PDE is discretized on a spatiotemporal grid and the resulting discrete set of equations is solved using numerical methods.\\ 
In turn, the number of the deployed computational elements and the structure of the PDE determine whether all of its scales have been resolved or whether the discretization amounts to a coarse-grained representation of the PDE. In the first case, the \textbf{fine grid simulation (FGS)} provides the discretized solution $\boldsymbol{\psi}$, whereas in \textbf{coarse grid simulation (CGS)} the resulting approximation is denoted by $\tilde{\boldsymbol{\psi}}$.\footnote{ In the following, all variables with the $\;\tilde {}\;$ are referring to the coarse-grid description.} The  RL policy can improve the accuracy of the solution $\tilde{\boldsymbol{\psi}}$ by introducing an appropriate  forcing term in the right-hand side of the CGS. For this purpose, FGS of the PDE are used as training episodes and serve as the ground truth to facilitate a reward signal. The CGS and FGS employed herein are introduced in the next section. The  proposed  grid-based RL framework is introduced in \cref{sec:RL}. 

%We introduce a Reinforcement Learning (RL) environment tailored to tackle the complexities of discovering closure models for dynamical systems. The environment integrates two distinct simulations:
%an Consider solving a partial differential equation (PDE) of the state variables $\psi$ and independent variables $C$ given certain initial conditions (ICs). The goal is to find the solution $\psi$ that statisfies $\text{PDE}(\psi, C) = 0 \text{ on } \Omega$, where $\Omega$ represents the domain. 

\subsection{Coarse and Fine Grid Simulation}
We consider a FGS of a spatiotemporal  PDE on a Cartesian  3D grid with temporal resolution $\Delta t$ for $N_T$ temporal steps with spatial resolution $\Delta x, \Delta y , \Delta z$ resulting in  $N_x, N_y, N_z$ discretization points. 
The CGS entails a coarser spatial discretization $  \widetilde{\Delta x}= d \Delta x$, $ \widetilde{\Delta y}= d \Delta y$, $ \widetilde{\Delta z}= d \Delta z$ as well as a coarser temporal discretization $ \widetilde{\Delta t} = d_t \Delta t$. Here, $d$ is the spatial and $d_t$ the temporal scaling factor. Consequently, at a time-step $\tilde{n}$, we define the discretized solution function of the CGS as $\tilde{\boldsymbol{\psi}} ^{\tilde{n}} \in \tilde \Psi := \mathbb R^{k \times \tilde N_x \times \tilde N_y \times \tilde N_z}$ with $k$ being the number of solution variables. The corresponding solution function of the FGS at time-step $n_f$ is $\boldsymbol{\psi}^{n_f} \in  \Psi := \mathbb R^{k \times N_x \times N_y \times N_z}$. The discretized solution function of the CGS can thus be described as a subsampled version of the FGS solution function and the subsampling operator $\mathcal S:  \Psi \rightarrow \tilde \Psi $ connects the two.\\The time stepping operator of the CGS $\mathcal G: \tilde \Psi \times \tilde{\mathcal C} \rightarrow \tilde \Psi $ leads to the update rule 
\begin{equation}
\label{eq:cg_dyn}
\tilde{\boldsymbol{\psi}}^{{\tilde{n}}+1} =\mathcal G(\tilde{\boldsymbol{\psi}}^{\tilde{n}}, \tilde C^{\tilde{n}}).
\end{equation}
Similarly, we define the time stepping operator of the FGS as $\mathcal F:  \Psi \times \mathcal C \rightarrow  \Psi $. In the case of the CGS,  $\tilde C \in \tilde{\mathcal{C}}$, is an adapted version of $C$, which for instance involves evaluating a PDE-parameter function using the coarse grid.\\
To simplify the notation, we set $n:={\tilde{n}}=d_t n_f$ in the following. %Hence, we apply $\mathcal{F}$ $d_T$-times in order to describe the same time instant with the same index $n$ in both CGS and FGS. 
The update rule of the FGS, i.e. applying $\mathcal{F}$ $d_T$-times to advance for the same time increment then the CGS,  is referred to as 
\begin{equation}
\boldsymbol{\psi}^{n+1} =\mathcal F^{d_t}(\boldsymbol{\psi}^n, C^n). 
\end{equation}
In \cref{sec:adintro} and \cref{sec:burgers} we are introducing the FGS and CGS for our illustrative examples. In line with our experiments and to further simplify notation, we are dropping the third spatial dimension in the following presentation.

\subsection{RL Environment}
\label{sec:RL}
\begin{figure}[ht]
\vskip 0.2in
\begin{center}
\centerline{\includegraphics[scale=0.075]{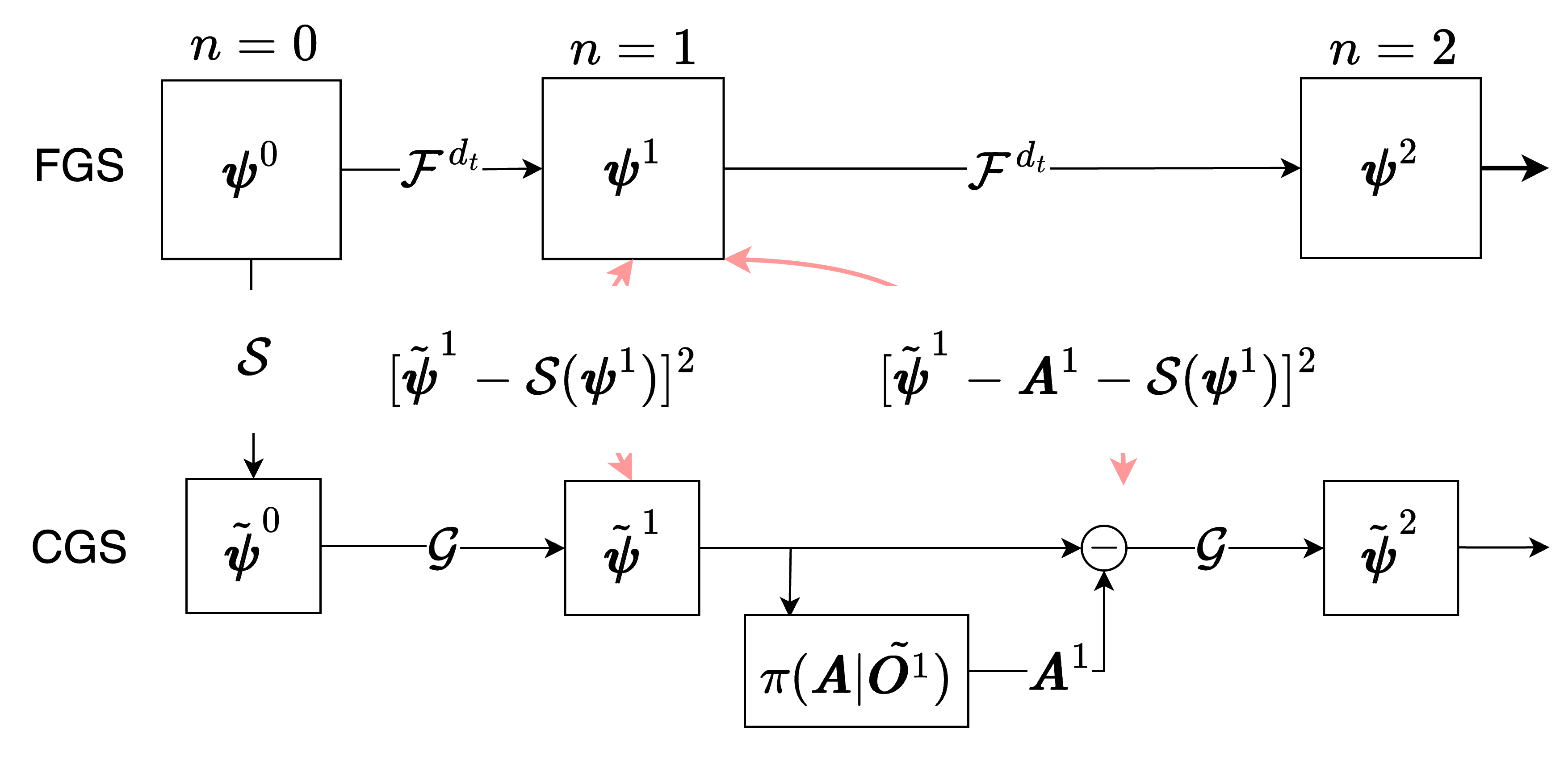}}
\caption{Illustration of the training environment with the agents embedded in the CGS. The reward measures how much the action taken improves the CGS.}
\label{fig:environment}
\end{center}
\vskip -0.2in
\end{figure}

The environment of our RL framework is summarized in \cref{fig:environment}. We define the state at step $n$ of the RL environment as the tuple $\boldsymbol{S}^n := (\boldsymbol{\psi}^n, \tilde{ \boldsymbol{\psi}^n})$. This state is only partially observable as the policy is acting only in the CGS. The observation $\boldsymbol{O}^n:=(\tilde{ \boldsymbol{\psi}^n}, \tilde C^n) \in \mathcal O$ is defined as the coarse representation of the discretized solution function and the PDEPs. Our goal is to train a policy $\pi$ that makes the dynamics of the CGS to be close to the dynamics of  the FGS. To achieve this goal, the action ${\boldsymbol{ A}^n} \in \mathcal A :=  \mathbb{R}^{k \times \tilde N_x \times \tilde N_y }$ at step $n$ of the environment is a collection of forcing terms for each discretization point of the CGS. In case the policy is later used to complement the CGS simulation the update function in \cref{eq:cg_dyn} changes to 
\begin{equation}
   \tilde{\boldsymbol{\psi}}^{n+1}=\mathcal G(\tilde{\boldsymbol{ \psi}}^n - {\boldsymbol{ A}^n}, \tilde C^n). 
\end{equation}
To encourage learning a policy that represents the non-resolved spatio-temporal scales, the reward is based on the difference between the CGS and FGS at time step $n$. In more detail, we define a local reward $\boldsymbol{R}^n \in \mathbb R^{\tilde N_x\times \tilde N_y}$ inspired by the reward proposed for image reconstruction in \cite{cv_pixel_rl}:
\begin{equation}
R_{ij}^n = \frac{1}{k} \sum_{w=1}^k \left([\tilde{\boldsymbol{ \psi}^n} - \mathcal S(\boldsymbol{\psi}^n)]^2 - [\tilde{\boldsymbol{ \psi}^{n}} - {\boldsymbol{ A}^n} - \mathcal S(\boldsymbol{\psi}^n)]^2 \right)_{wij}
\end{equation}
Here, the square $[\cdot]^2$ is computed per matrix entry. We note that the reward function therefore returns a matrix that gives a scalar reward for each discretization point of the CGS.
If the action $\boldsymbol{A}^n$ is bringing the discretized solution function of the CGS $\tilde{\boldsymbol{ \psi}^n}$ closer to the subsampled discretized solution function of the FGS $\mathcal{S}(\boldsymbol{ \psi}^n)$, the reward is positive, and vice versa. In case $\boldsymbol{A}^n$ corresponds to the zero matrix, the reward is the zero matrix as well.\\
During the training process, the objective is to find the optimal policy $\pi^*$ that maximizes the mean expected reward per discretization point given the discount rate $\gamma$ and the observations:
\begin{equation}
\label{eq:pol_old}
\pi^*  =\underset{\pi}{\operatorname{argmax }}\; \mathbb E_{\pi}\left(\sum_{n=0}^{\infty} \gamma^n \bar{r}^{n}\right)
\end{equation}
with the mean expected reward
\begin{equation}
\bar{r}^{n}  =\frac{1}{\tilde N_x\cdot \tilde N_y} \sum_{i,j=1}^{\tilde N_x, \tilde N_y}  R_{ij}^{n}.
\end{equation}
%To summarize, the environment is set up in such a way that it incentivizes the agents to keep the CGS and FGS close w.r.t. the discretization point-wise squared error. In order to fulfill this task successfully, the agents need to learn to balance compensating for the deficiencies of the CGS simulation, which arise due to the coarser grid and discretization scheme, while also keeping the simulation stable. 

\subsection{Grid-based RL Formulation} 
\label{sec:marl}
The policy $\pi$ predicts a local action $ \boldsymbol{A}^n_{i,j} \in \mathbb R^k$ at each discretization point which implies a very high dimensional  continuous action space. Hence, formulating the closures  with a single agent is very challenging. However, since the rewards are designed to be inherently local, locality can be used as inductive bias and the RL learning framework can be interpreted as a multi-agent problem \cite{cv_pixel_rl}. One agent is placed at each discretization point of the coarse grid with a  corresponding local reward $R_{ij}^n$. We remark  that this approach augments adaptivity as  one can place extra agents at additional, suitably selected, discretization points.\\
Each agent develops its own optimal policy, which we later defined to be shared, and \cref{eq:pol_old} is replaced by 
\begin{equation} \label{eq:CNN-MARL_objective}
\pi_{ij}^*=\underset{\pi_{ij}}{\operatorname{argmax}} \ \mathbb E_{\pi_{ij}}\left(\sum_{n=0}^{\infty} \gamma^n  R_{ij}^{n}\right),
\end{equation}
Here, we used $\mathcal O (\tilde N_x \tilde N_y)$ agents, which for typically used  grid sizes of numerical simulations, becomes a large number compared to typical MARL problem settings \cite{albrecht2023multi}.\\
We parametrize the local policies using neural networks. However, since training this many individual neural nets can become  computationally expensive, we parametrize all the agents together using  one fully convolutional network (FCN) \cite{fcn}, which implies that the agents share one policy.

\subsection{Parallelizing Local Policies with a FCN}  
All local policies are parametrized using one FCN such that one forward pass through the FCN computes the forward pass for all the agents at once. This approach enforces the aforementioned locality and the receptive field of the FCN corresponds to the spatial neighborhood that an agent at a given discretization point can observe.\\
We define the collection of policies in the FCN as $\Pi^{FCN}: \mathcal O  \rightarrow \mathcal{A}$.
In further discussions, we will refer to $\Pi^{FCN}$ as the \textit{global policy}. The policy $\pi_{ij}$ of the agent at point $(i,j)$ is subsequently implicitly defined through the global policy as 
    $\pi_{ij}( O_{ij}) := \left [ \Pi^{FCN} (\boldsymbol O)\right ]_{:ij}.$
Here, $ \boldsymbol{O}_{ij}$ contains only a part of the input contained in $\boldsymbol O$ \footnote{The exact content of $\boldsymbol{O}_{ij}$ is depending on the receptive field of the FCN}. For consistency, we will refer to $\boldsymbol O_{ij}$ as a \textit{local observation} and $\boldsymbol O$ as the \textit{global observation}. We note that the policies $\pi_{ij}(\boldsymbol{O}_{ij})$ map the observation to a probability distribution at each discretization point (see \cref{sec:nn} for details including the neural network architecure employed).
Similar to the global policy, the global value function is parametrized using a FCN as well. It maps the global observation to an expected return $H \in  \mathcal{H
} :=R^{ \tilde N_x \times \tilde N_y}$ at each discretization point
$\mathcal V^{FCN}: \mathcal O \rightarrow \mathcal{H}.$
Similarly to the local policies, we define the \textit{local value function} related to the agent at point $(i,j)$ as $v_{ ij}( O_{ij}) :=  [ \mathcal V^{FCN}(\boldsymbol O) ]_{ij}.$

\subsection{Policy Optimization } \label{sec:algorithmic_details}
In order to solve the  optimization problem in \cref{eq:CNN-MARL_objective}, we employ a modified version of the PPO algorithm \cite{ppo}.\\
Policy updates are performed by taking gradient steps on
\begin{equation}
\mathbb{E}_{\boldsymbol S, \boldsymbol A \sim \Pi^{FCN}} \left[ \frac{1}{\tilde N_x\cdot \tilde N_y} \sum_{i,j=1}^{\tilde N_x, \tilde N_y} L_{\pi_{ij}}(\boldsymbol{O}_{ij},  \boldsymbol{A}_{ij}) \right]
\end{equation}
with the local version of the PPO objective $L_{\pi_{ij}}( \boldsymbol{O}_{ij},  \boldsymbol{A}_{ij})$. This corresponds to the local objective of the policy of the agent at point $(i,j)$ \cite{ppo}.\\
The global value function is trained on the MSE loss
\[ L_{\mathcal{V}}(\boldsymbol O^n, \boldsymbol{G}^n) =  ||\mathcal V^{FCN}(\boldsymbol O^n)- \boldsymbol{G}^n||_2^2 \]
where $\boldsymbol{G}^n\in \mathbb R^{\tilde N_x\times \tilde N_y}$ represents the actual global return observed by interaction with the environment and is computed as $\boldsymbol{G}^n=\sum_{i=n}^N \gamma^{i-n} \boldsymbol{R}^i$. Here, $N$ represents the length of the respective trajectory.\\
We have provided an overview of the modified PPO algorithm in \cref{appendix:ppo} together with further details regarding the local version of the PPO objective $L_{\pi_{ij}}( \boldsymbol{O}_{ij},  \boldsymbol{A}_{ij})$. Our implementation of the adapted PPO algorithm is based on the single agent PPO algorithm of the Tianshou framework \cite{tianshou}.

\subsection{Computational Complexity}
\label{sec:computational}
We note that the computational complexity of the CGS w.r.t. the number of discretization points scales with $\mathcal O(\tilde N_x \tilde N_y)$. As one forward pass through the FCN also scales with $\mathcal O(\tilde N_x \tilde N_y)$ the same is true for Closure-RL. The FGS employs a finer grid, which leads to a computational cost that scales with $\mathcal O(d_t d ^2 \tilde N_x \tilde N_y)$. This indicates a scaling advantage for Closure-RL compared to a FGS. Based on these considerations and as shown in the experiments, Closure-RL is able to compress some of the computations that are performed on the fine grid as it is able to significantly improve the CGS while keeping the execution time below that of the FGS.

\section{Experiments}
\label{sec:exp}
\subsection{Advection Equation}
\label{sec:adintro}
First we apply Closure-RL to the 2D advection equation\footnote{The code to reproduce all results in this paper can be found at (URL added upon publication).}:
\begin{equation}
\frac{\partial \psi}{\partial t} + u(x,y)\frac{\partial \psi}{\partial x} + v(x,y)\frac{\partial \psi}{\partial y} = 0,
\end{equation}
where $\psi$ represents a physical concentration that is transported by the velocity field $C=(u, v)$. We employ periodic boundary conditions (PBCs) on the domain \(\Omega = [0, 1] \times [0, 1]\).

For the FGS, this domain is discretized using $N_x=N_y=256$ discretization points in each dimension. To guarantee stability, we employ a time step that ensures that the Courant–Friedrichs–Lewy (CFL) \cite{Courant_1928} condition is fulfilled. The spatial derivatives are calculated using central differences and the time stepping uses the fourth-order Runge-Kutta scheme \cite{quarteroni2008numerical}. The FGS is fourth order accurate in time and second order accurate in space.

We construct the CGS by employing the subsampling factors $d=4$ and $d_t=4$.  Spatial derivatives in the CGS use an  upwind scheme and time stepping is performed with the forward Euler method, resulting in first order accuracy in both space and time. All settings of numerical values used for the CGS and FGS are summarized in \cref{tab:advection_params}.

\subsubsection{Initial Conditions}
In order to prevent overfitting and promote  generalization, we design the initializations of $\psi, u$ and $v$ to be different for each episode while still fulfilling the PBCs and guaranteeing the incompressibility of the velocity field. 
The velocity fields are sampled from a distribution $\mathcal D^{Vortex}_{Train}$ by taking a linear combination of Taylor-Greene vortices and an additional random translational field. Further details are provided in \cref{sec:train_vortices}. For visualization purposes, the concentration of a new episode is set to a random sample from the MNIST dataset \cite{mnist} that is scaled to values in the range $[0, 1]$. In order to increase the diversity of the initializations, we augment the data by performing random rotations of $\pm 90 ^{\circ}$ in the image loading pipeline.
\begin{figure*} 
\centering
    \subfigure{\includegraphics[width=0.24\textwidth]{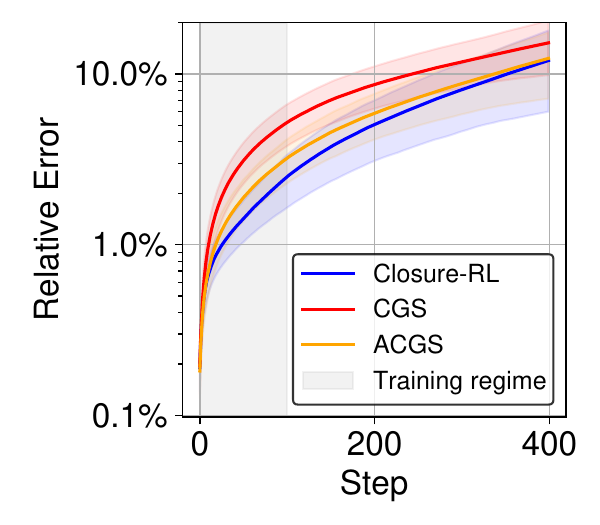}}
    \subfigure{\includegraphics[width=0.24\textwidth]{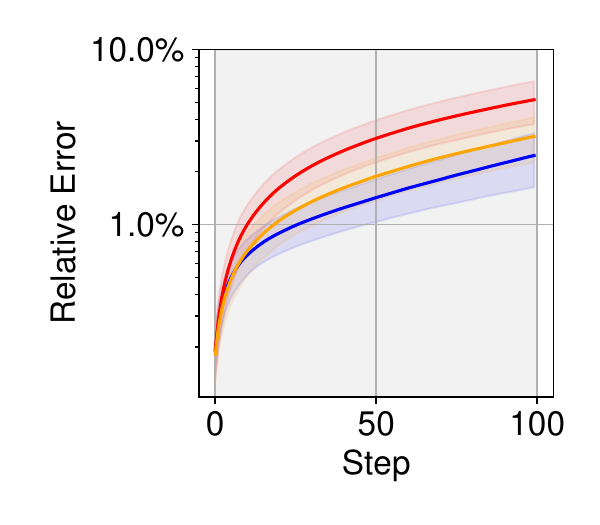}}
    \subfigure{\includegraphics[width=0.24\textwidth]{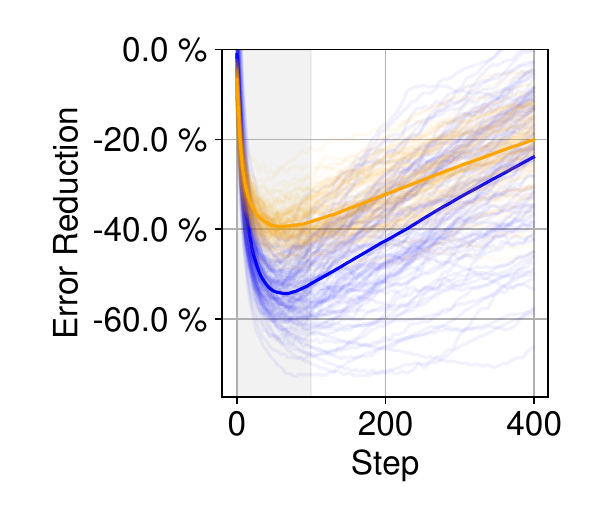}}
    \subfigure{\includegraphics[width=0.24\textwidth]{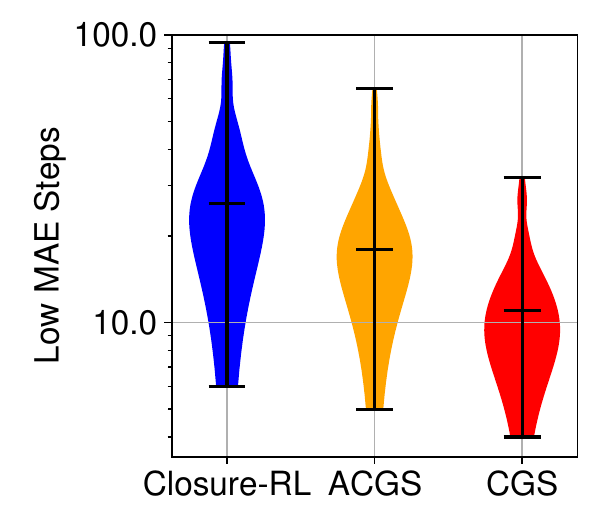}}
    \caption{Results of Closure-RL applied to the advection equation. The relative MAEs are computed over 100 simulations with respect to (w.r.t.) the FGS. The shaded regions correspond to the respective standard deviations. The violin plot on the right shows the number of simulation steps until the relative MAE w.r.t. the FGS reaches the threshold of $1\%$.}
    \label{fig:advection_rollout}
\end{figure*}

\subsubsection{Training}
\label{Sec:train}
We train the framework for a total of 2000 epochs and collect 1000 transitions in each epoch. More details regarding the training process as well as the training time  are provided in \cref{sec:training_hyperparams}.\\
We note that the amount and length of episodes varies during the training process:
The episodes are truncated based on the relative Mean Absolute Error ($\text{MAE}$) defined as $\text{MAE}(  \psi^n, \tilde{ \psi^n}) = \frac{1}{\tilde N_x\cdot \tilde N_y}\sum_{i,j} |\psi^n_{ij}- \tilde{\psi}^n_{ij}| / \psi_{max}$
between the CGS and FGS concentrations. Here, $\psi_{max}$ is the maximum observable value of the concentration $\psi$. If this error exceeds the threshold of $1.5 \%$, the episode is truncated. This ensures that during training, the CGS and FGS stay close to each other so that the reward signal is meaningful. As the agents become better during the training process, the mean episode length increases as the two simulations stay closer to each other for longer. We designed this adaptive training procedure in order to obtain stable simulations.\\
We note that inspired by \cite{novati2021automating} we experimented with adding an additional domain-specific global term to the reward (see \cref{sec:glo} for details). We did not see any improvements compared to our Closure-RL formulation which indicates that the proposed reward structure is sufficient to learn an accurate closure model.

\subsubsection{Accurate Coarse Grid Simulation}
We also introduce the \textit{'accurate coarse grid simulation'} (ACGS) in order to further compare the effecst of numerics and grid size in CGS and FGS. ACGS operates on the coarse grid, just like the CGS, with a higher order numerical scheme, than the CGS, so that it has the same order of accuracy as the FGS.

\subsubsection{In- and Out-of-Distribution MAE}
\begin{figure}[ht]
\vskip 0.2in
\begin{center}
\centerline{\includegraphics[width=0.78\columnwidth]{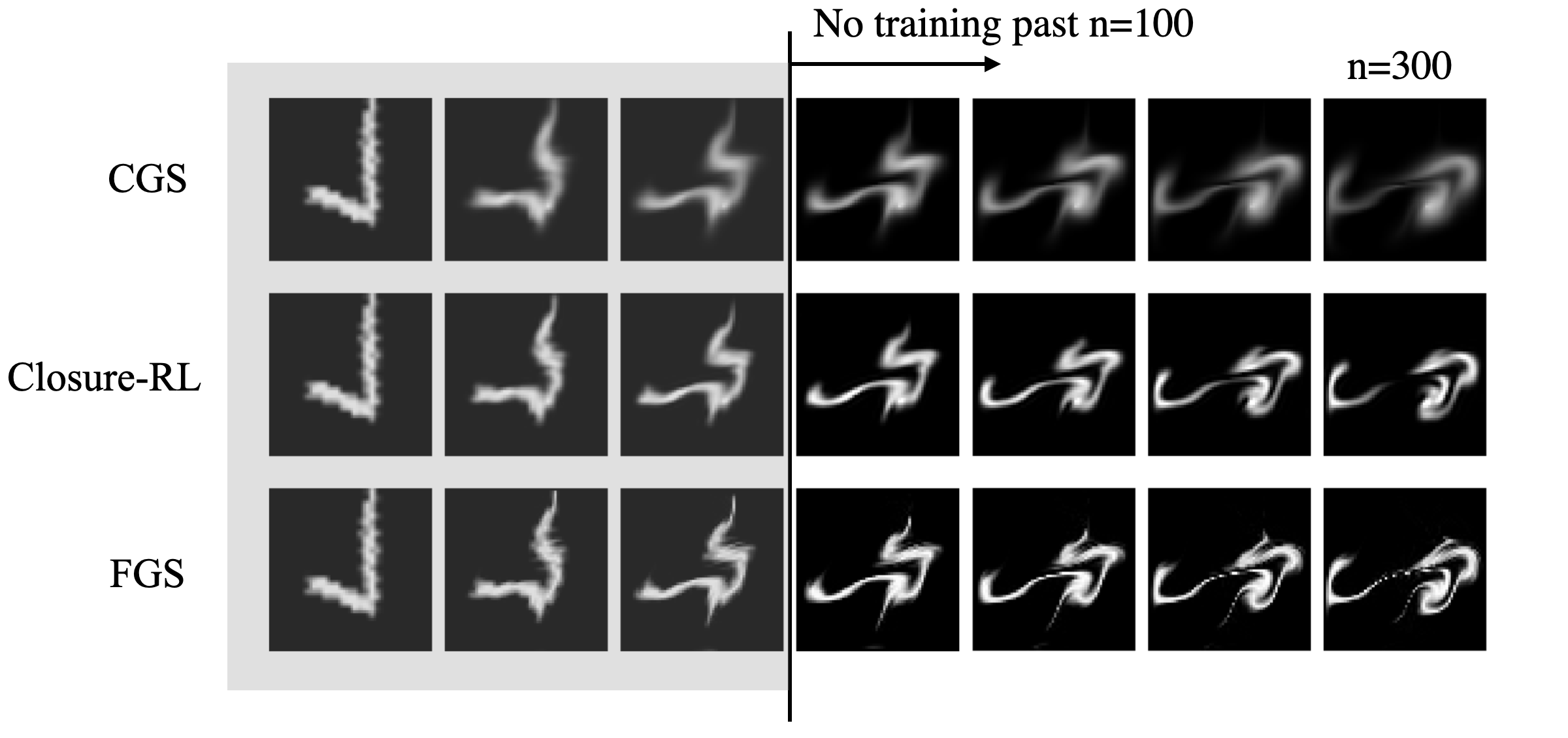}}
\caption{Example run comparing CGS, Closure-RL and FGS with the same ICs. The concentration is sampled from the test set and the velocity components are randomly sampled from $\mathcal D^{Vortex}_{Train}$. Note that the agents of the Closure-RLmethod have only been trained up to $n=100$. However, they qualitatively improve the CGS simulation past that point.}
\label{fig:advection_example}
\end{center}
\vskip -0.2in
\end{figure}
We develop  metrics for Closure-RL by running 100 simulations of 50 time steps each with different ICs. For the in-distribution case, the concentrations are sampled from the MNIST test set and the velocity fields are sampled from $\mathcal D^{Vortex}_{Train}$. To quantify the performance on out-of-distribution ICs, we also run evaluations on simulations using the Fashion-MNIST dataset (F-MNIST) \cite{fashionmnist} and a new distribution, $\mathcal D^{Vortex}_{Test}$, for the velocity fields. The latter is defined in \cref{sec:test_vortices}. The resulting error metrics of CGS, ACGS and Closure-RL w.r.t. the FGS are collected in \cref{tab:50_step_mae_advection}. A qualitative example of the CGS, FGS and the Closure-RL method after training is presented in \cref{fig:advection_example}. The example shows that Closure-RL is able to compensate for the dissipation  that is introduced by the first order scheme and coarse grid in the CGS.

\begin{table}[!ht]
\caption{Relative MAE at time step 50 averaged over 100 simulations with different ICs to both the velocity and concentration. All relative MAE values are averaged over the complete domain and reported in percent.}
\label{tab:50_step_mae_advection}
\vskip 0.15in
\centering
\scriptsize 
\begin{tabular}{@{}l|cccr@{}}
\toprule
Velocity & \multicolumn{2}{c}{$\mathcal D^{Vortex}_{Train}$} & \multicolumn{2}{c}{$\mathcal D^{Vortex}_{Test}$} \\
Concentr. & MNIST & F-MNIST & MNIST & F-MNIST \\
\hline 
CGS & $3.13 \pm 0.80$ & $3.23 \pm 0.92$ & $3.82 \pm 0.78$ & $3.12 \pm 0.73$ \\ 
ACGS & $1.90 \pm 0.51$ & $2.27 \pm 0.64$ & $2.28 \pm 0.47$ & $2.23 \pm 0.50$\\
Closure-RL & $\textbf{1.46} \pm 0.33$ & $\textbf{2.12} \pm 0.57$  & $\textbf{1.58} \pm 0.37$ & $\textbf{2.04} \pm 0.56$ \\  \bottomrule
\multicolumn{5}{c}{Relative Improvements w.r.t. CGS} \\
 \bottomrule
ACGS &  -39\% &  -30\% &  -40\% &  -31\%\\
 Closure-RL & \textbf{-53\%} & \textbf{-34\%} & \textbf{-58\%} & \textbf{-36\%} \\ 
 \bottomrule
\end{tabular}
\end{table}
Closure-RL reduces the relative MAE of the CGS after 50 steps by $30\%$ or more in both in- and out-of-distribution cases. This shows that the agents have learned a meaningful correction for the truncation errors of the numerical schemes in the coarser grid. Closure-RL also outperforms the ACGS w.r.t. the MAE metric which indicates that the learned corrections  emulate a higher-order scheme. This indicates that the proposed methodology is  able to emulate the unresolved dynamics and is a suitable option for complementing existing numerical schemes.

\textbf{We note the strong performance of our framework w.r.t. the out-of-distribution examples.} For both unseen and out-of-distribution ICs as well as PDEPs, the framework was able to outperform CGS and ACGS. In our opinion, this indicates that we have discovered an actual model of the forcing terms that goes beyond the training scenarios. 

\subsubsection{Evolution of Numerical Error}
The results in the previous sections are mostly focused on the difference between the methods after a rollout of 50 time steps. To analyze how the methods compare over the course of a longer rollout, we analyze the relative MAE at each successive step of a simulation with MNIST and $\mathcal D^{Vortex}_{Train}$ as distributions for the ICs. The results are shown in \cref{fig:advection_rollout}.\\ 
The plots of the evolution of the relative error show that Closure-RL is able to improve the CGS for the entire range of a 400-step rollout, although it has only been trained for 100 steps. This implies that the agents are seeing distributions of the concentration that have never been encountered during training and are able to generalize to these scenarios. When measuring the duration of simulations for which the relative error stays below $1\%$, we observe that the Closure-RL method outperforms both ACGS and CGS, indicating that the method is able to produce simulations with higher long term stability than CGS and ACGS. We attribute this to our adaptive scheme for episode truncation during training as introduced in \cref{Sec:train} and note that the increased stability can be observed well beyond the training regime.

\subsubsection{Interpretation of Actions}
The intention of our closure scheme is to negate the errors introduced by the numerical scheme. For the advection CGS, we are able to show that the actions taken are indeed highly correlated with the optimal action based on the errors of the numerical scheme used. In \cref{sec:int}, we are reporting the derivation of this optimal action as well as the correlation with the actions actually taken by Closure-RL. The obtained correlation coefficients are between $0.7-0.82$ depending on the task.

\subsection{Burgers' Equation}
\label{sec:burgers}

\begin{figure*}[h]
\centering
    \subfigure{\includegraphics[width=0.24\textwidth]{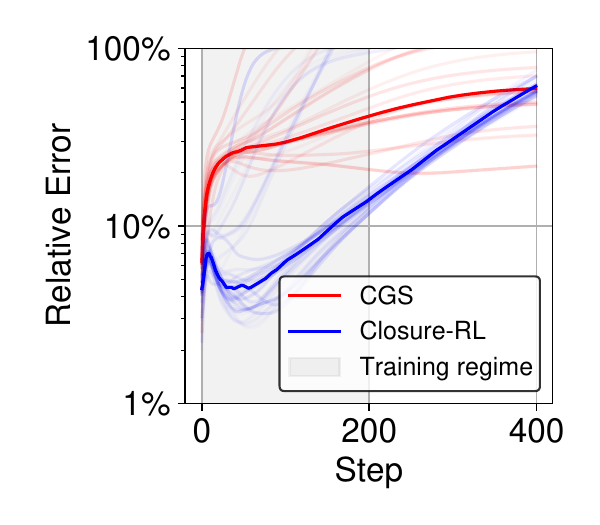}}
    \subfigure{\includegraphics[width=0.24\textwidth]{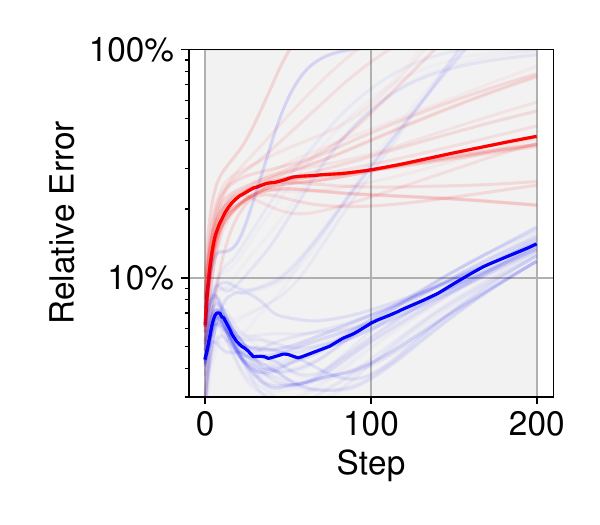}}
    \subfigure{\includegraphics[width=0.24\textwidth]{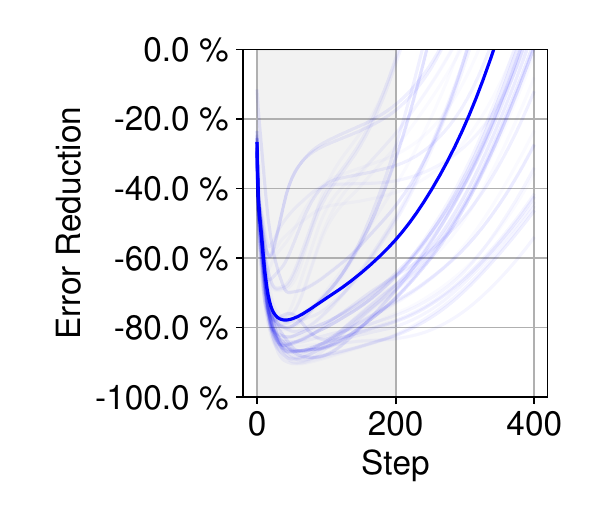}}
    \subfigure{\includegraphics[width=0.24\textwidth]{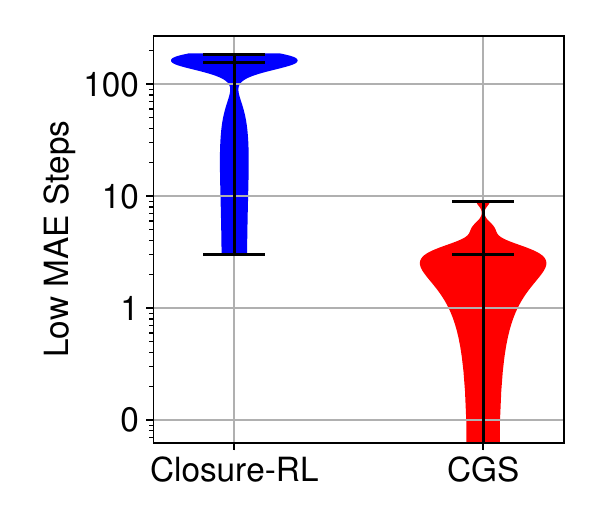}}
    \caption{Results of Closure-RL applied to the Burgers' equation. The relative MAEs are computed over 100 simulations with respect to the FGS. The shaded regions correspond to the respective standard deviations. The violin plot on the right shows the number of simulation steps until the relative MAE w.r.t. the FGS reaches the threshold of $10\%$.}
    \label{fig:burgers_errors}
\end{figure*}
As a second example, we apply our framework to the 2D viscous Burgers' equation:
\begin{equation}
\frac{\partial \boldsymbol{\psi}}{\partial t} + (\boldsymbol{\psi} \cdot \nabla) \boldsymbol{\psi} - \nu \nabla^2 \boldsymbol{\psi} = 0.
\end{equation}
Here, $\boldsymbol{\psi} := (u, v)$ consists of both velocity components and the PDE has the single input PDEP $C=\nu$. As for the advection equation, we assume  periodic boundary conditions on the domain \(\Omega = [0, 1] \times [0, 1]\). 
In comparison to the advection example, we are now dealing with two solution variables and thus $k=2$. 

For the FGS, the aforementioned domain is discretized using $N_x=N_y=150$ discretization points in each dimension. Moreover, we again choose $\Delta t$ to fulfill the CFL condition for stability (see \cref{tab:burgers_params}). The spatial derivatives are calculated using the upwind scheme and the forward Euler method is used for the time stepping \cite{quarteroni2008numerical}.
We construct the CGS by employing the subsampling factors $d=5$ and $d_t=10$. For the Burgers experiment, we apply a mean filter $K$ with kernel size $d\times d$ before the actual subsampling operation. The mean filter is used to eliminate higher frequencies in the fine grid state variables, which would lead to accumulating high errors. The CGS employs the same numerical schemes as the FGS here. This leads to first order accuracy in both space and time. All numerical settings used for the CGS and FGS are collected in \cref{tab:burgers_params}. In this example, where FGS and CGS are using the same numerical scheme, the Closure-RL framework has to focus solely on negating the effects of the coarser discretization. 
For training and evaluation, we generate random, incompressible velocity fields as ICs (see \cref{sec:train_vortices} for details) and set the viscosity $\nu$ to $0.003$. We observed that the training of the predictions actions can be improved by multiplying the predicted forcing terms with $\widetilde{\Delta t}$. This is consistent with our previous analysis in \cref{sec:int} as the optimal action is also multiplied but this factor. Again, we train the model for 2000 epochs with 1000 transitions each. The maximum episode length during training is set to $200$ steps and, again, we truncate the episodes adaptively, when the relative error exceeds $20 \%$. Further details on training Closure-RL for the Burgers' equation can be found in \cref{sec:training_hyperparams}. Visualization of the results can be found in \cref{sec:adres} and the resulting relative errors w.r.t. the FGS are shown in \cref{fig:burgers_errors}. The Closure-RL method again improves the CGS significantly, also past the point of the 200 steps seen during training. Specifically, in the range of step $0$ to step $100$ in which the velocity field changes fastest, we see a significant error reductions up to $-80\%$. When analyzing the duration for which the episodes stay under the relative error of $10\%$, we observe that the mean number of steps is improved by two order of magnitude, indicating that the method is able to improve the long term accuracy of the CGS.  

\section{Conclusions}
We propose Closure-RL, a novel framework for the automated discovery of closure models of coarse grained discretizations of  time-dependent PDEs. It utilizes a grid-based RL formulation with a FCN for both the policy and the value network. This enables the incorporation of local rewards without necessitating individual neural networks for each agent and allows to efficiently train a large number of agents. Moreover, the framework trains on rollouts without needing to backpropagate through the numerical solver itself.
We show that Closure-RL develops a  policy that compensates for numerical errors in a CGS of both the 2D advection and Burgers' equation. More importantly we find that the learned closure model can be used for predictions in extrapolative test cases. 

Further work may focus on extending the formulation to irregular grids by using a Graph Convolutional Network \cite{DBLP:journals/corr/KipfW16} instead of the FCN. Similarly to the current approach, we can place one agent at each discretization point whose receptive field now depends on the connection between the graph nodes. Moreover, for very large systems of interest the numerical schemes used are often multi-grid and the grid-based RL framework should reflect this. For such cases, we suggest defining separate rewards for each of the grids employed by the numerical solver. Additionally, Closure-RL may serve as a stepping-stone towards incorporating further inductive bias or constraints as its action space can readily be adapted. For instance, the actions for some agents could be explicitly parameterized or coupled with their neighboring agents. In this regard, Closure-RL could also be used to further improve other, domain-specific closure models.  

\section*{Acknowledgements}
S.K. and P.K. acknowledge support by The European High Performance Computing Joint Undertaking (EuroHPC) Grant DCoMEX
(956201-H2020-JTI-EuroHPC-2019-1) and support by the Defense Advanced Research Projects Agency (DARPA) through Award
HR00112490489.

% Reference
% For natbib users:
\bibliography{references,references2}

\begin{thebibliography}{54}
\providecommand{\natexlab}[1]{#1}
\providecommand{\url}[1]{\texttt{#1}}
\expandafter\ifx\csname urlstyle\endcsname\relax
  \providecommand{\doi}[1]{doi: #1}\else
  \providecommand{\doi}{doi: \begingroup \urlstyle{rm}\Url}\fi

\bibitem[Wilcox(1988)]{wilcox1988multiscale}
David~C Wilcox.
\newblock Multiscale model for turbulent flows.
\newblock \emph{AIAA journal}, 26\penalty0 (11):\penalty0 1311--1320, 1988.

\bibitem[Dura-Bernal et~al.(2019)Dura-Bernal, Suter, Gleeson, Cantarelli, Quintana, Rodriguez, Kedziora, Chadderdon, Kerr, Neymotin, et~al.]{dura2019netpyne}
Salvador Dura-Bernal, Benjamin~A Suter, Padraig Gleeson, Matteo Cantarelli, Adrian Quintana, Facundo Rodriguez, David~J Kedziora, George~L Chadderdon, Cliff~C Kerr, Samuel~A Neymotin, et~al.
\newblock Netpyne, a tool for data-driven multiscale modeling of brain circuits.
\newblock \emph{Elife}, 8:\penalty0 e44494, 2019.

\bibitem[Council(2012)]{climatenas}
National~Research Council.
\newblock \emph{A National Strategy for Advancing Climate Modeling}.
\newblock The National Academies Press, 2012.

\bibitem[Mahadevan(2016)]{mahadevan2016impact}
Amala Mahadevan.
\newblock The impact of submesoscale physics on primary productivity of plankton.
\newblock \emph{Annual review of marine science}, 8:\penalty0 161--184, 2016.

\bibitem[Hey(2009)]{hey2009fourth}
Tony Hey.
\newblock \emph{The fourth paradigm}.
\newblock United States of America., 2009.

\bibitem[Rossinelli et~al.(2013)Rossinelli, Hejazialhosseini, Hadjidoukas, Bekas, Curioni, Bertsch, Futral, Schmidt, Adams, and Koumoutsakos]{rossinelli2013a}
Diego Rossinelli, Babak Hejazialhosseini, Panagiotis Hadjidoukas, Costas Bekas, Alessandro Curioni, Adam Bertsch, Scott Futral, Steffen~J. Schmidt, Nikolaus~A. Adams, and Petros Koumoutsakos.
\newblock 11 {PFLOP/s} simulations of cloud cavitation collapse.
\newblock In \emph{Proceedings of the International Conference on High Performance Computing, Networking, Storage and Analysis}, SC '13, pages 3:1--3:13, New York, NY, USA, 2013. ACM.
\newblock ISBN 978-1-4503-2378-9.
\newblock \doi{10.1145/2503210.2504565}.
\newblock URL \url{http://doi.acm.org/10.1145/2503210.2504565}.

\bibitem[Peng et~al.(2021)Peng, Alber, Buganza~Tepole, Cannon, De, Dura-Bernal, Garikipati, Karniadakis, Lytton, Perdikaris, et~al.]{peng2021multiscale}
Grace~CY Peng, Mark Alber, Adrian Buganza~Tepole, William~R Cannon, Suvranu De, Savador Dura-Bernal, Krishna Garikipati, George Karniadakis, William~W Lytton, Paris Perdikaris, et~al.
\newblock Multiscale modeling meets machine learning: What can we learn?
\newblock \emph{Archives of Computational Methods in Engineering}, 28:\penalty0 1017--1037, 2021.

\bibitem[Sanderse et~al.(2024)Sanderse, Stinis, Maulik, and Ahmed]{sanderse2024scientific}
Benjamin Sanderse, Panos Stinis, Romit Maulik, and Shady~E. Ahmed.
\newblock Scientific machine learning for closure models in multiscale problems: a review, 2024.

\bibitem[Furuta et~al.(2019)Furuta, Inoue, and Yamasaki]{cv_pixel_rl}
Ryosuke Furuta, Naoto Inoue, and Toshihiko Yamasaki.
\newblock Pixelrl: Fully convolutional network with reinforcement learning for image processing, 2019.

\bibitem[Bergdorf et~al.(2005)Bergdorf, Cottet, and Koumoutsakos]{bergdorf2005multilevel}
Michael Bergdorf, Georges-Henri Cottet, and Petros Koumoutsakos.
\newblock Multilevel adaptive particle methods for convection-diffusion equations.
\newblock \emph{Multiscale Modeling \& Simulation}, 4\penalty0 (1):\penalty0 328--357, 2005.

\bibitem[Raissi et~al.(2019)Raissi, Perdikaris, and Karniadakis]{raissi2019physics}
Maziar Raissi, Paris Perdikaris, and George~E Karniadakis.
\newblock Physics-informed neural networks: A deep learning framework for solving forward and inverse problems involving nonlinear partial differential equations.
\newblock \emph{Journal of Computational physics}, 378:\penalty0 686--707, 2019.

\bibitem[Karniadakis et~al.(2021)Karniadakis, Kevrekidis, Lu, Perdikaris, Wang, and Yang]{karniadakis2021physics}
George~Em Karniadakis, Ioannis~G Kevrekidis, Lu~Lu, Paris Perdikaris, Sifan Wang, and Liu Yang.
\newblock Physics-informed machine learning.
\newblock \emph{Nature Reviews Physics}, 3\penalty0 (6):\penalty0 422--440, 2021.

\bibitem[Lu et~al.(2021)Lu, Jin, Pang, Zhang, and Karniadakis]{lu2021learning}
Lu~Lu, Pengzhan Jin, Guofei Pang, Zhongqiang Zhang, and George~Em Karniadakis.
\newblock Learning nonlinear operators via deeponet based on the universal approximation theorem of operators.
\newblock \emph{Nature machine intelligence}, 3\penalty0 (3):\penalty0 218--229, 2021.

\bibitem[Li et~al.(2020)Li, Kovachki, Azizzadenesheli, Liu, Bhattacharya, Stuart, and Anandkumar]{li2020fourier}
Zongyi Li, Nikola Kovachki, Kamyar Azizzadenesheli, Burigede Liu, Kaushik Bhattacharya, Andrew Stuart, and Anima Anandkumar.
\newblock Fourier neural operator for parametric partial differential equations.
\newblock \emph{arXiv preprint arXiv:2010.08895}, 2020.

\bibitem[Seidman et~al.(2022)Seidman, Kissas, Perdikaris, and Pappas]{seidman2022nomad}
Jacob Seidman, Georgios Kissas, Paris Perdikaris, and George~J Pappas.
\newblock Nomad: Nonlinear manifold decoders for operator learning.
\newblock \emph{Advances in Neural Information Processing Systems}, 35:\penalty0 5601--5613, 2022.

\bibitem[Brandstetter et~al.(2023)Brandstetter, Berg, Welling, and Gupta]{brandstetter2022clifford}
Johannes Brandstetter, Rianne van~den Berg, Max Welling, and Jayesh~K Gupta.
\newblock Clifford neural layers for pde modeling.
\newblock \emph{ICLR}, 2023.

\bibitem[Kaltenbach et~al.(2023)Kaltenbach, Perdikaris, and Koutsourelakis]{kaltenbach2023semi}
Sebastian Kaltenbach, Paris Perdikaris, and Phaedon-Stelios Koutsourelakis.
\newblock Semi-supervised invertible neural operators for bayesian inverse problems.
\newblock \emph{Computational Mechanics}, pages 1--20, 2023.

\bibitem[Karnakov et~al.(2022)Karnakov, Litvinov, and Koumoutsakos]{karnakov2022optimizing}
Petr Karnakov, Sergey Litvinov, and Petros Koumoutsakos.
\newblock Optimizing a discrete loss (odil) to solve forward and inverse problems for partial differential equations using machine learning tools.
\newblock \emph{arXiv preprint arXiv:2205.04611}, 2022.

\bibitem[Yang and Wang(2020)]{yang2020overview}
Yaodong Yang and Jun Wang.
\newblock An overview of multi-agent reinforcement learning from game theoretical perspective.
\newblock \emph{arXiv preprint arXiv:2011.00583}, 2020.

\bibitem[Freed et~al.(2021)Freed, Kapoor, Abraham, Schneider, and Choset]{freed2021learning}
Benjamin Freed, Aditya Kapoor, Ian Abraham, Jeff Schneider, and Howie Choset.
\newblock Learning cooperative multi-agent policies with partial reward decoupling.
\newblock \emph{IEEE Robotics and Automation Letters}, 7\penalty0 (2):\penalty0 890--897, 2021.

\bibitem[Wen et~al.(2022)Wen, Kuba, Lin, Zhang, Wen, Wang, and Yang]{wen2022multi}
Muning Wen, Jakub Kuba, Runji Lin, Weinan Zhang, Ying Wen, Jun Wang, and Yaodong Yang.
\newblock Multi-agent reinforcement learning is a sequence modeling problem.
\newblock \emph{Advances in Neural Information Processing Systems}, 35:\penalty0 16509--16521, 2022.

\bibitem[Albrecht et~al.(2023)Albrecht, Christianos, and Sch{\"a}fer]{albrecht2023multi}
Stefano~V Albrecht, Filippos Christianos, and Lukas Sch{\"a}fer.
\newblock Multi-agent reinforcement learning: Foundations and modern approaches.
\newblock \emph{Massachusetts Institute of Technology: Cambridge, MA, USA}, 2023.

\bibitem[Sutton et~al.(2023)Sutton, Machado, Holland, Szepesvari, Timbers, Tanner, and White]{sutton2023reward}
Richard~S Sutton, Marlos~C Machado, G~Zacharias Holland, David Szepesvari, Finbarr Timbers, Brian Tanner, and Adam White.
\newblock Reward-respecting subtasks for model-based reinforcement learning.
\newblock \emph{Artificial Intelligence}, 324:\penalty0 104001, 2023.

\bibitem[Novati et~al.(2021)Novati, de~Laroussilhe, and Koumoutsakos]{novati2021automating}
Guido Novati, Hugues~Lascombes de~Laroussilhe, and Petros Koumoutsakos.
\newblock Automating turbulence modelling by multi-agent reinforcement learning.
\newblock \emph{Nature Machine Intelligence}, 3\penalty0 (1):\penalty0 87--96, 2021.

\bibitem[Bae and Koumoutsakos(2022)]{bae2022scientific}
H~Jane Bae and Petros Koumoutsakos.
\newblock Scientific multi-agent reinforcement learning for wall-models of turbulent flows.
\newblock \emph{Nature Communications}, 13\penalty0 (1):\penalty0 1443, 2022.

\bibitem[Ling et~al.(2016)Ling, Kurzawski, and Templeton]{ling2016reynolds}
Julia Ling, Andrew Kurzawski, and Jeremy Templeton.
\newblock Reynolds averaged turbulence modelling using deep neural networks with embedded invariance.
\newblock \emph{Journal of Fluid Mechanics}, 807:\penalty0 155--166, 2016.

\bibitem[Durbin(2018)]{durbin2018some}
Paul~A Durbin.
\newblock Some recent developments in turbulence closure modeling.
\newblock \emph{Annual Review of Fluid Mechanics}, 50:\penalty0 77--103, 2018.

\bibitem[Lippe et~al.(2023)Lippe, Veeling, Perdikaris, Turner, and Brandstetter]{lippe2023pde}
Phillip Lippe, Bastiaan~S Veeling, Paris Perdikaris, Richard~E Turner, and Johannes Brandstetter.
\newblock Pde-refiner: Achieving accurate long rollouts with neural pde solvers.
\newblock \emph{arXiv preprint arXiv:2308.05732}, 2023.

\bibitem[Gupta and Brandstetter(2022)]{gupta2022towards}
Jayesh~K Gupta and Johannes Brandstetter.
\newblock Towards multi-spatiotemporal-scale generalized pde modeling.
\newblock \emph{arXiv preprint arXiv:2209.15616}, 2022.

\bibitem[Goyal and Bengio(2022)]{goyal2022inductive}
Anirudh Goyal and Yoshua Bengio.
\newblock Inductive biases for deep learning of higher-level cognition.
\newblock \emph{Proceedings of the Royal Society A}, 478\penalty0 (2266):\penalty0 20210068, 2022.

\bibitem[Kaltenbach and Koutsourelakis(2020)]{kaltenbach2020incorporating}
Sebastian Kaltenbach and Phaedon-Stelios Koutsourelakis.
\newblock Incorporating physical constraints in a deep probabilistic machine learning framework for coarse-graining dynamical systems.
\newblock \emph{Journal of Computational Physics}, 419:\penalty0 109673, 2020.

\bibitem[Yin et~al.(2021)Yin, Le~Guen, Dona, de~B{\'e}zenac, Ayed, Thome, and Gallinari]{yin2021augmenting}
Yuan Yin, Vincent Le~Guen, J{\'e}r{\'e}mie Dona, Emmanuel de~B{\'e}zenac, Ibrahim Ayed, Nicolas Thome, and Patrick Gallinari.
\newblock Augmenting physical models with deep networks for complex dynamics forecasting.
\newblock \emph{Journal of Statistical Mechanics: Theory and Experiment}, 2021\penalty0 (12):\penalty0 124012, 2021.

\bibitem[Wang et~al.(2021)Wang, Wang, and Perdikaris]{wang2021learning}
Sifan Wang, Hanwen Wang, and Paris Perdikaris.
\newblock Learning the solution operator of parametric partial differential equations with physics-informed deeponets.
\newblock \emph{Science advances}, 7\penalty0 (40):\penalty0 eabi8605, 2021.

\bibitem[Wang et~al.(2022{\natexlab{a}})Wang, Sankaran, and Perdikaris]{wang2022respecting}
Sifan Wang, Shyam Sankaran, and Paris Perdikaris.
\newblock Respecting causality is all you need for training physics-informed neural networks.
\newblock \emph{arXiv preprint arXiv:2203.07404}, 2022{\natexlab{a}}.

\bibitem[Greydanus et~al.(2019)Greydanus, Dzamba, and Yosinski]{greydanus2019hamiltonian}
Samuel Greydanus, Misko Dzamba, and Jason Yosinski.
\newblock Hamiltonian neural networks.
\newblock \emph{Advances in neural information processing systems}, 32, 2019.

\bibitem[Kaltenbach and Koutsourelakis(2021)]{kaltenbach2021physics}
Sebastian Kaltenbach and Phaedon-Stelios Koutsourelakis.
\newblock Physics-aware, probabilistic model order reduction with guaranteed stability.
\newblock \emph{ICLR}, 2021.

\bibitem[Cranmer et~al.(2020)Cranmer, Greydanus, Hoyer, Battaglia, Spergel, and Ho]{cranmer2020lagrangian}
Miles Cranmer, Sam Greydanus, Stephan Hoyer, Peter Battaglia, David Spergel, and Shirley Ho.
\newblock Lagrangian neural networks.
\newblock \emph{arXiv preprint arXiv:2003.04630}, 2020.

\bibitem[Uchendu et~al.(2023)Uchendu, Xiao, Lu, Zhu, Yan, Simon, Bennice, Fu, Ma, Jiao, et~al.]{uchendu2023jump}
Ikechukwu Uchendu, Ted Xiao, Yao Lu, Banghua Zhu, Mengyuan Yan, Jos{\'e}phine Simon, Matthew Bennice, Chuyuan Fu, Cong Ma, Jiantao Jiao, et~al.
\newblock Jump-start reinforcement learning.
\newblock In \emph{International Conference on Machine Learning}, pages 34556--34583. PMLR, 2023.

\bibitem[Walke et~al.(2023)Walke, Yang, Yu, Kumar, Orbik, Singh, and Levine]{walke2023don}
Homer~Rich Walke, Jonathan~Heewon Yang, Albert Yu, Aviral Kumar, J{\k{e}}drzej Orbik, Avi Singh, and Sergey Levine.
\newblock Don’t start from scratch: Leveraging prior data to automate robotic reinforcement learning.
\newblock In \emph{Conference on Robot Learning}, pages 1652--1662. PMLR, 2023.

\bibitem[Long et~al.(2015)Long, Shelhamer, and Darrell]{fcn}
Jonathan Long, Evan Shelhamer, and Trevor Darrell.
\newblock Fully convolutional networks for semantic segmentation, 2015.

\bibitem[Schulman et~al.(2017)Schulman, Wolski, Dhariwal, Radford, and Klimov]{ppo}
John Schulman, Filip Wolski, Prafulla Dhariwal, Alec Radford, and Oleg Klimov.
\newblock Proximal policy optimization algorithms.
\newblock \emph{CoRR}, abs/1707.06347, 2017.
\newblock URL \url{http://arxiv.org/abs/1707.06347}.

\bibitem[Weng et~al.(2022)Weng, Chen, Yan, You, Duburcq, Zhang, Su, Su, and Zhu]{tianshou}
Jiayi Weng, Huayu Chen, Dong Yan, Kaichao You, Alexis Duburcq, Minghao Zhang, Yi~Su, Hang Su, and Jun Zhu.
\newblock Tianshou: A highly modularized deep reinforcement learning library.
\newblock \emph{Journal of Machine Learning Research}, 23\penalty0 (267):\penalty0 1--6, 2022.
\newblock URL \url{http://jmlr.org/papers/v23/21-1127.html}.

\bibitem[Courant et~al.(1928)Courant, Friedrichs, and Lewy]{Courant_1928}
R.~Courant, K.~Friedrichs, and H.~Lewy.
\newblock \"{U}ber die partiellen differenzengleichungen der mathematischen physik.
\newblock \emph{Mathematische Annalen}, 100\penalty0 (1):\penalty0 32--74, December 1928.
\newblock \doi{10.1007/BF01448839}.
\newblock URL \url{http://dx.doi.org/10.1007/BF01448839}.

\bibitem[Quarteroni and Valli(2008)]{quarteroni2008numerical}
Alfio Quarteroni and Alberto Valli.
\newblock \emph{Numerical approximation of partial differential equations}, volume~23.
\newblock Springer Science \& Business Media, 2008.

\bibitem[Deng(2012)]{mnist}
Li~Deng.
\newblock The mnist database of handwritten digit images for machine learning research.
\newblock \emph{IEEE Signal Processing Magazine}, 29\penalty0 (6):\penalty0 141--142, 2012.

\bibitem[Xiao et~al.(2017)Xiao, Rasul, and Vollgraf]{fashionmnist}
Han Xiao, Kashif Rasul, and Roland Vollgraf.
\newblock Fashion-mnist: a novel image dataset for benchmarking machine learning algorithms, 2017.

\bibitem[Kipf and Welling(2016)]{DBLP:journals/corr/KipfW16}
Thomas~N. Kipf and Max Welling.
\newblock Semi-supervised classification with graph convolutional networks.
\newblock \emph{CoRR}, abs/1609.02907, 2016.
\newblock URL \url{http://arxiv.org/abs/1609.02907}.

\bibitem[Zhang et~al.(2017)Zhang, Zuo, Gu, and Zhang]{ircnn}
Kai Zhang, Wangmeng Zuo, Shuhang Gu, and Lei Zhang.
\newblock Learning deep cnn denoiser prior for image restoration, 2017.

\bibitem[Schulman et~al.(2018)Schulman, Moritz, Levine, Jordan, and Abbeel]{generalized_advantage_estimation}
John Schulman, Philipp Moritz, Sergey Levine, Michael Jordan, and Pieter Abbeel.
\newblock High-dimensional continuous control using generalized advantage estimation, 2018.

\bibitem[Khorashadizadeh et~al.(2024)Khorashadizadeh, Liaudat, Liu, McEwen, and Dokmanic]{khorashadizadehlofi}
AmirEhsan Khorashadizadeh, Tob{\i}as~I Liaudat, Tianlin Liu, Jason~D McEwen, and Ivan Dokmanic.
\newblock Lofi: Neural local fields for scalable image reconstruction.
\newblock \emph{arXiv preprint arXiv:2411.04995}, 2024.

\bibitem[Ronneberger et~al.(2015)Ronneberger, Fischer, and Brox]{ronneberger2015u}
Olaf Ronneberger, Philipp Fischer, and Thomas Brox.
\newblock U-net: Convolutional networks for biomedical image segmentation.
\newblock In \emph{Medical image computing and computer-assisted intervention--MICCAI 2015: 18th international conference, Munich, Germany, October 5-9, 2015, proceedings, part III 18}, pages 234--241. Springer, 2015.

\bibitem[Liu et~al.(2021)Liu, Lin, Cao, Hu, Wei, Zhang, Lin, and Guo]{liu2021swin}
Ze~Liu, Yutong Lin, Yue Cao, Han Hu, Yixuan Wei, Zheng Zhang, Stephen Lin, and Baining Guo.
\newblock Swin transformer: Hierarchical vision transformer using shifted windows.
\newblock In \emph{Proceedings of the IEEE/CVF international conference on computer vision}, pages 10012--10022, 2021.

\bibitem[Liang et~al.(2021)Liang, Cao, Sun, Zhang, Van~Gool, and Timofte]{liang2021swinir}
Jingyun Liang, Jiezhang Cao, Guolei Sun, Kai Zhang, Luc Van~Gool, and Radu Timofte.
\newblock Swinir: Image restoration using swin transformer.
\newblock In \emph{Proceedings of the IEEE/CVF international conference on computer vision}, pages 1833--1844, 2021.

\bibitem[Wang et~al.(2022{\natexlab{b}})Wang, Cun, Bao, Zhou, Liu, and Li]{wang2022uformer}
Zhendong Wang, Xiaodong Cun, Jianmin Bao, Wengang Zhou, Jianzhuang Liu, and Houqiang Li.
\newblock Uformer: A general u-shaped transformer for image restoration.
\newblock In \emph{Proceedings of the IEEE/CVF conference on computer vision and pattern recognition}, pages 17683--17693, 2022{\natexlab{b}}.

\end{thebibliography}

%%%%%%%%%%%%%%%%%%%%%%%%%%%%%%%%%%%%%%%%%%%%%%%%%%%%%%%%%%%%

\appendix
%\section{Appendix}

\section{Neural Network Architecture}
\label{sec:nn}
\begin{figure}[ht]
\vskip 0.2in
\begin{center}
\centerline{\includegraphics[scale=0.075]{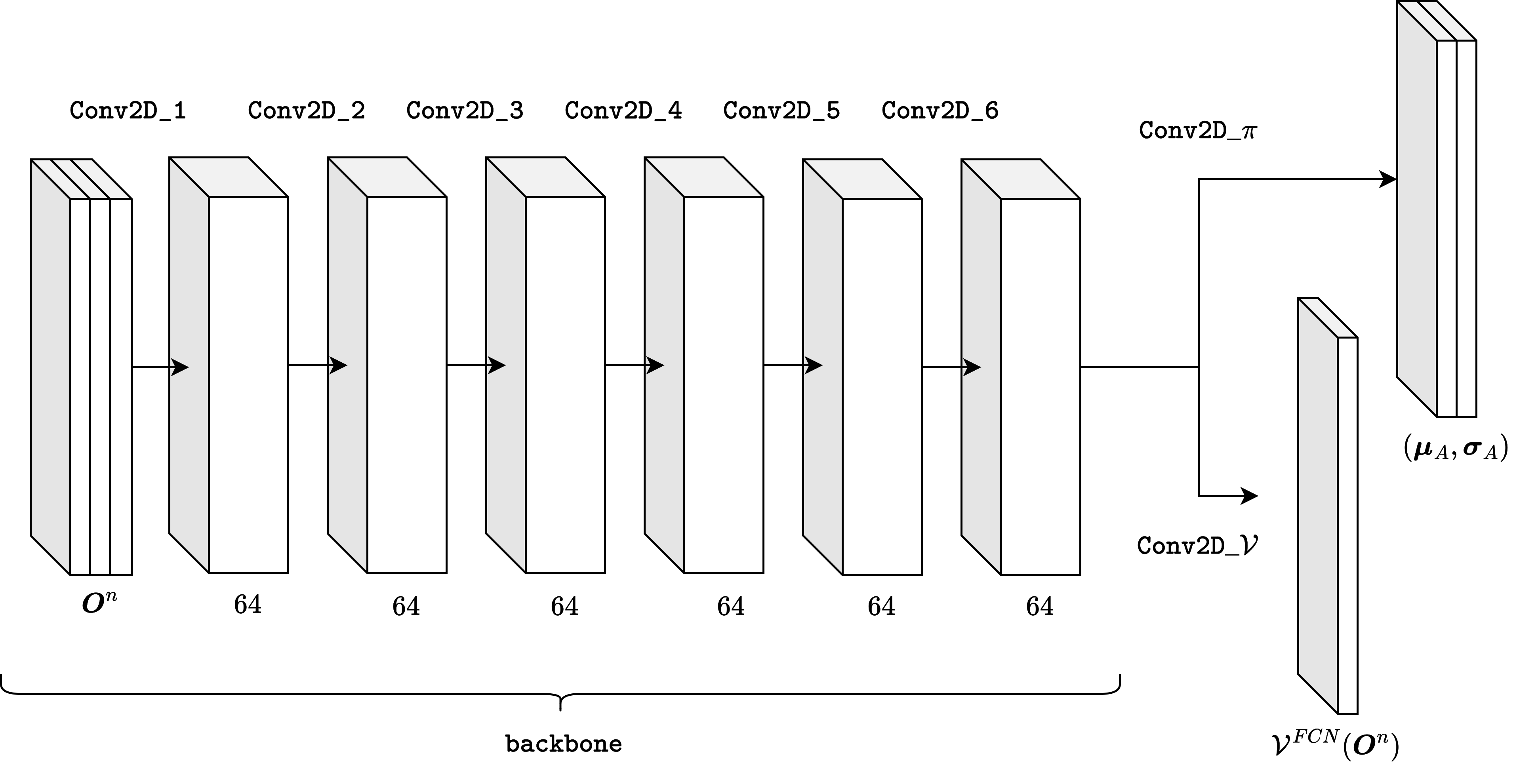}}
\caption{The IRCNN backbone takes the current global observation and maps it to a feature tensor. This feature tensor is passed into two different convolutional layers that predict the per-discretization point action-distribution-parameters and state-values. In the case of a PDE with three spatial dimensions, the architecture would need to be based on three-dimensional convolutional layers instead. }
\label{fig:ircnn}
\end{center}
\vskip -0.2in
\end{figure}
The optimal policy is expected to compensate for errors introduced by the numerical method and its implementation on a  coarse grid. We use the Image Restoration CNN (IRCNN) architecture proposed in \cite{ircnn} as our backbone for the policy- and value-network. A discussion of this choice can be found in \cref{sec:choice}. In \cref{fig:ircnn} we present an illustration of the architecture and show that the policy- and value network share the same backbone. The policy- and value network only differ by their last convolutional layer, which takes the features extracted by the backbone and maps them either to the action-distribution-parameters $\boldsymbol \mu_A \in \mathcal A$ and $\boldsymbol \sigma_A \in \mathcal A$ or the predicted return value for each agent.  The local policies are assumed to be independent, such that we can write the distribution at a specific discretization point as
\begin{equation}
\pi_{ij}( \boldsymbol{A}_{ij}| \boldsymbol{O}_{ij}) = \mathcal N( \boldsymbol{\mu}_{A,ij},  \boldsymbol{\sigma}_{A,ij}).
\end{equation}
During training, the actions are sampled to allow for exploration, and during inference only the mean is taken as the action of the agent.\\
We note that the padding method used for the FCN can incorporate boundary conditions into the architecture. For instance, in the case of periodic boundary conditions, we propose to use circular padding that involves wrapping around values from one end of the input tensor to the other.

\newpage
\section{Adapted PPO Algorithm} \label{appendix:ppo}
As defined in \cref{sec:algorithmic_details} the loss function for the global value function is 
\[ L_{\mathcal{V}}(\boldsymbol O^n, \boldsymbol{G}^n, \phi) =  ||\mathcal V^{FCN}_{\phi}(\boldsymbol O^n)- \boldsymbol{G}^n||_2^2. \]
We note that this notation contains the weights $\phi$, which parameterize the underlying neural network.

The objective for the global policy is defined as
\[L_{\Pi}(\boldsymbol O^n, \boldsymbol A^n, \theta_k, \theta, \beta) := \frac{1}{\tilde N_x\cdot \tilde N_y} \sum_{i,j=1}^{\tilde N_x, \tilde N_y} L_{\pi_{ij}}({\boldsymbol{O}}^n_{ij}, \boldsymbol{A}^n_{ij}, \theta_k, \theta) - \beta H(\pi_{ij,\theta}),\]
where  $H(\pi_{ij,\theta})$ is the entropy of the local policy and $L_{\pi_{ij}}$ is the standard single-agent PPO-Clip objective
\[ \label{eq:ppo_objective}
L_{\pi_{ij}}(o, a, \theta_k, \theta) = \min \left( \frac{\pi_{ij,\theta}(a|o)}{\pi_{ij,\theta_k}(a|o)} Adv^{\pi_{ij,\theta_k}}(o, a), \text{clip} \left( \frac{\pi_{ij,\theta}(a|o)}{\pi_{ij,\theta_k}(a|o)}, 1 - \epsilon, 1 + \epsilon \right) Adv^{\pi_{ij,\theta_k}}(o, a) \right).
\]
The advantage estimates \( {Adv}^{\pi_{ij,\theta_k}} \) are computed with generalized advantage estimation (GAE) \cite{generalized_advantage_estimation} using the output of the global value network \( \mathcal{V}^{FCN}_{\phi} \).\\

The resulting adapted PPO algorithm is presented in \cref{algo:ppo}. Major differences compared to the original PPO algorithm are vectorized versions of the value network loss and PPO-Clip objective, as well as a summation over all the discretization points of the domain before performing an update step.  

\begin{algorithm}[tb]
   \caption{Adapted PPO Algorithm}
    \label{algo:ppo}
\begin{algorithmic}
   \STATE {\bfseries Input:} Initial policy weights \( \theta_1 \), initial value function weights \( \phi_1 \), \\
 clip ratio \( \epsilon\), discount factor \(\gamma\), entropy regularization weight \(\beta\)
   \FOR{\( k = 1, 2, \ldots \)}
   \STATE Collect set of trajectories \( D_k  \) with the  global policy \( \Pi^{FCN}_{\theta_k} \)
   \STATE \textit{\# Update global policy}
   \STATE Update $\theta$ by performing a SGD step on\\ \( \theta_{k+1} = \arg \max_\theta \frac{1}{\|D_k\|}\sum_{\boldsymbol{O}^n, \boldsymbol A^n \in D_k} L_{\Pi}(\boldsymbol O^n, \boldsymbol A^n, \theta_k, \theta, \beta) \)
   \STATE \textit{\# Update global value network}
    \STATE Compute returns for each transition using \( \boldsymbol{G}^n=\sum_{i=n}^N \gamma^{i-t} \boldsymbol{R}^i \) where $N$ is the length of the respective trajectory
    \STATE Update $\phi$ by performing a SGD step on \( \phi_{k+1} = \arg \min_\phi \frac{1}{\|D_k\|}\sum_{\boldsymbol{O}^n, \boldsymbol G^n \in D_k} L_{\mathcal V}(\boldsymbol{O}^n, \boldsymbol{G}^n, \phi) \)
   \ENDFOR
\end{algorithmic}
\end{algorithm}

\newpage
\section{Additional Results}
\subsection{Advection Equation}
\begin{figure}[ht]
\vskip 0.2in
\begin{center}
\centerline{\includegraphics[width=0.75\columnwidth]{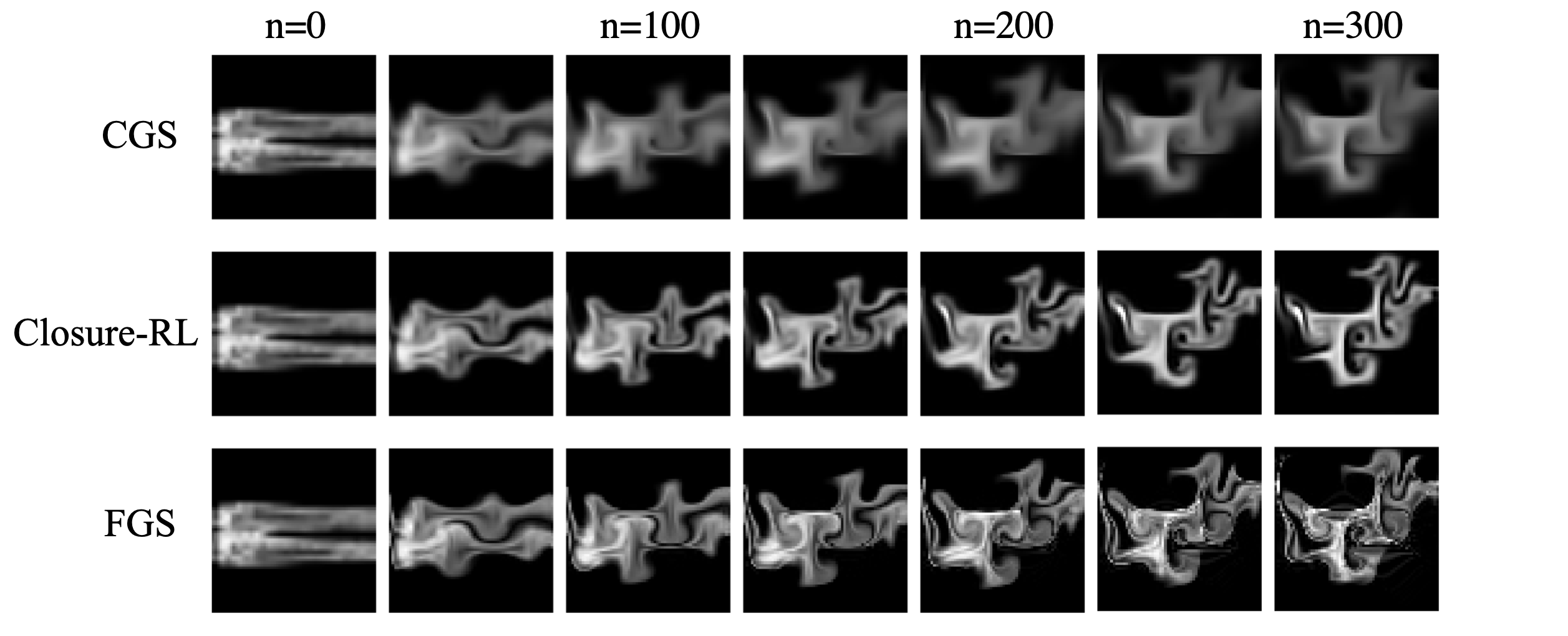}}
\caption{$\boldsymbol \psi^0$ is sampled from the MNIST test set. The velocity field is sampled from $\mathcal D^{Vortex}_{Test}$ (See \cref{sec:train_vortices}). Here, the IC of the concentration comes from a different distribution than the one used for training.}
\end{center}
\vskip -0.2in
\end{figure}

\begin{figure}[ht]
\vskip 0.2in
\begin{center}
\centerline{\includegraphics[width=0.75\columnwidth]{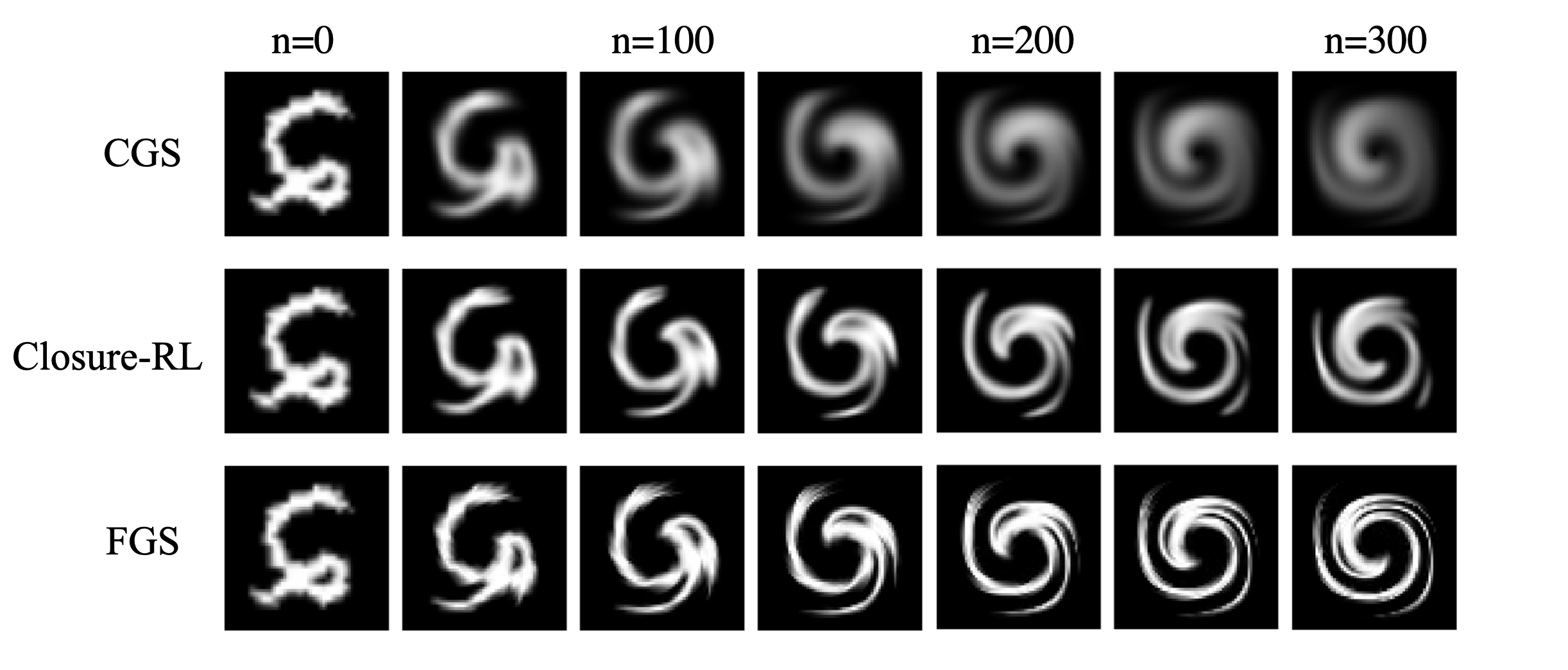}}
\caption{$\boldsymbol \psi^0$ is a sample from the F-MNIST test set. The velocity field is sampled from $\mathcal D^{Vortex}_{Train}$ (See \cref{sec:test_vortices}). Note that the velocity field comes from a different distribution than the one used for training.}
\end{center}
\vskip -0.2in
\end{figure}

\newpage
\subsection{Burgers' Equation}
\label{sec:adres}
\begin{figure}[ht]
\vskip 0.2in
\begin{center}
\centerline{\includegraphics[width=0.75\columnwidth]{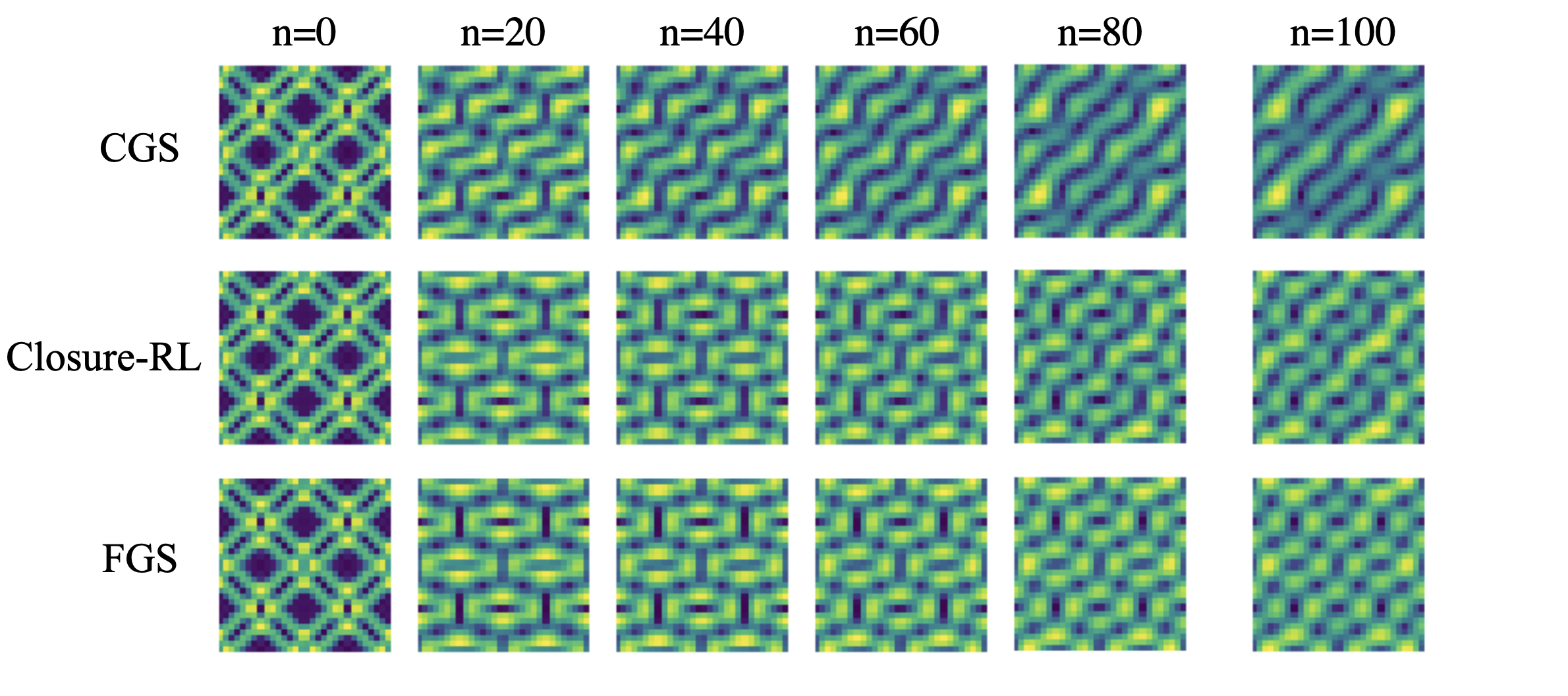}}
\caption{$\boldsymbol \psi^0$ is a sample from $\mathcal D_{Train}^{Vortex}$. We plot the velocity magnitude of every $20$th step of simulation with $100$ coarse time steps. The example shows that Closure-RL does also qualitatively keep the simulation closer to the FGS. The failure of the CGS to account for the subgrid-scale dynamics leads to diverging trajectories.}
\label{fig:burgers_example}
\end{center}
\vskip -0.2in
\end{figure}

\begin{figure}[ht]
\vskip 0.2in
\begin{center}
\centerline{\includegraphics[width=0.75\columnwidth]{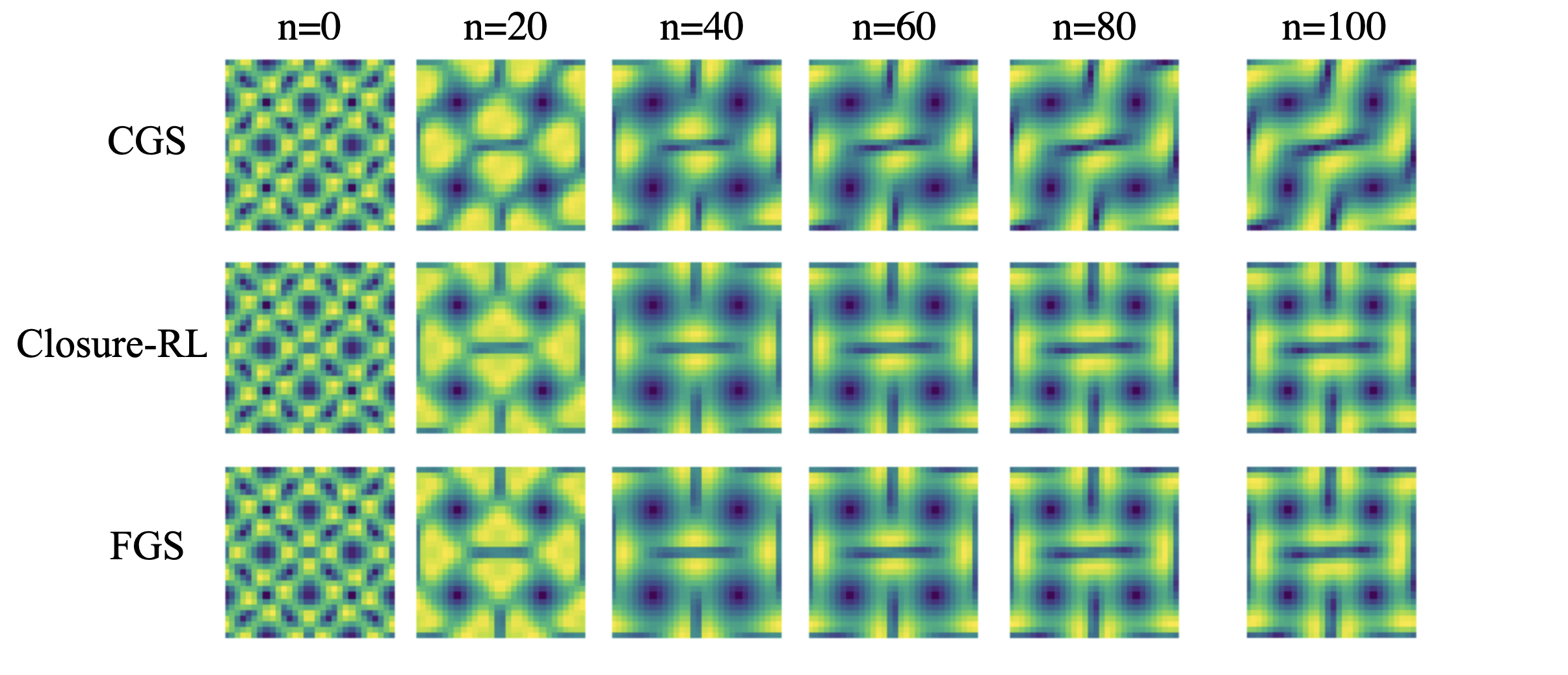}}
\caption{Velocity magnitude plotted for a 100-step roll out. The set-up is the same as in \cref{fig:burgers_example} and the example again shows, that Closure-RL leads to qualitatively different dynamics during a roll-out that are closer to those of the FGS.}
\end{center}
\vskip -0.2in
\end{figure}

\newpage
\section{Technical Details on Hyperparamters and Training Runs} \label{sec:training_hyperparams}
\begin{figure}[h]
    \centering
    \hfill
    \raisebox{-\height}{\includegraphics[width=0.4\linewidth]{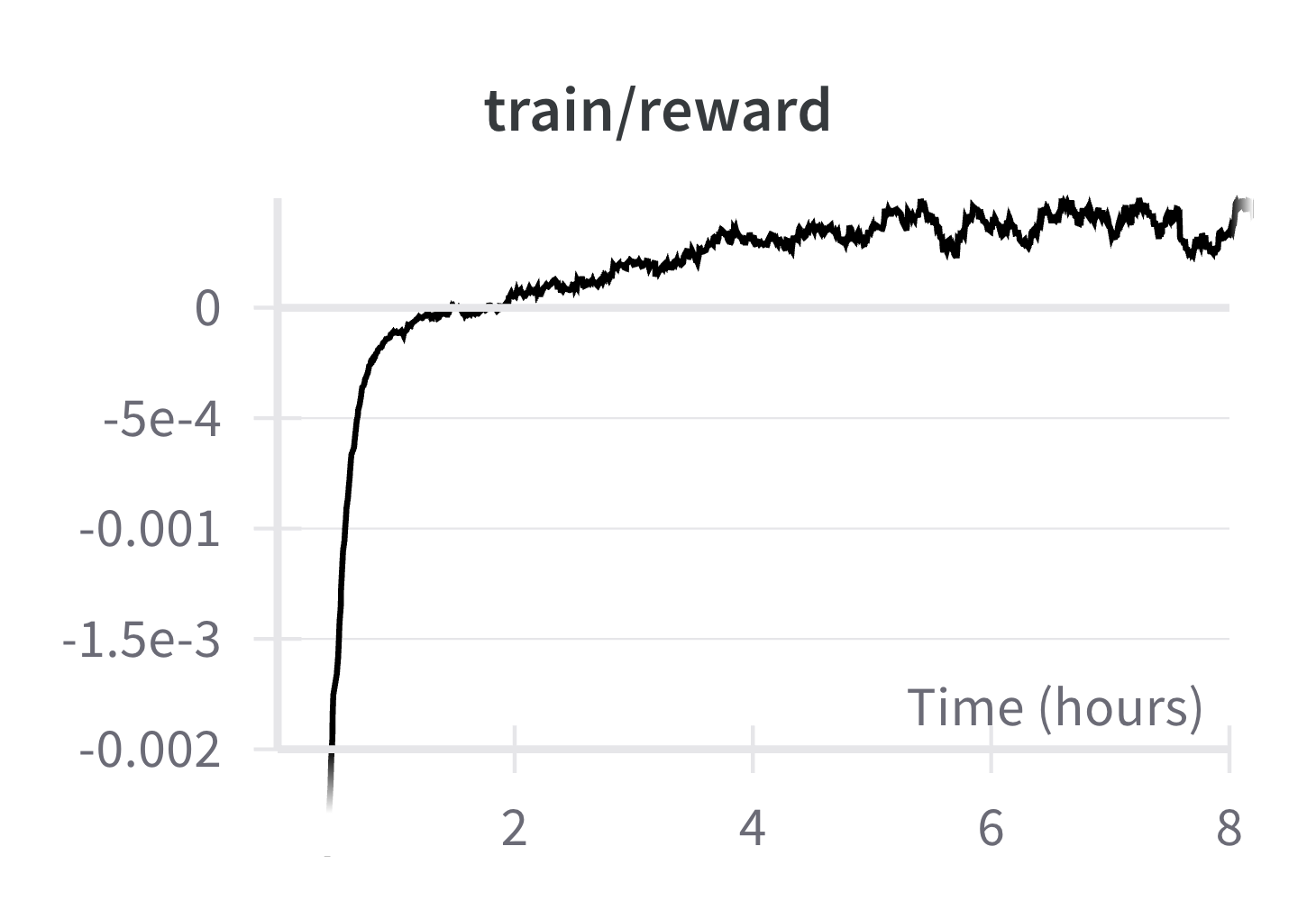}}
    \hfill
    \raisebox{-\height}{\includegraphics[width=0.4\linewidth]{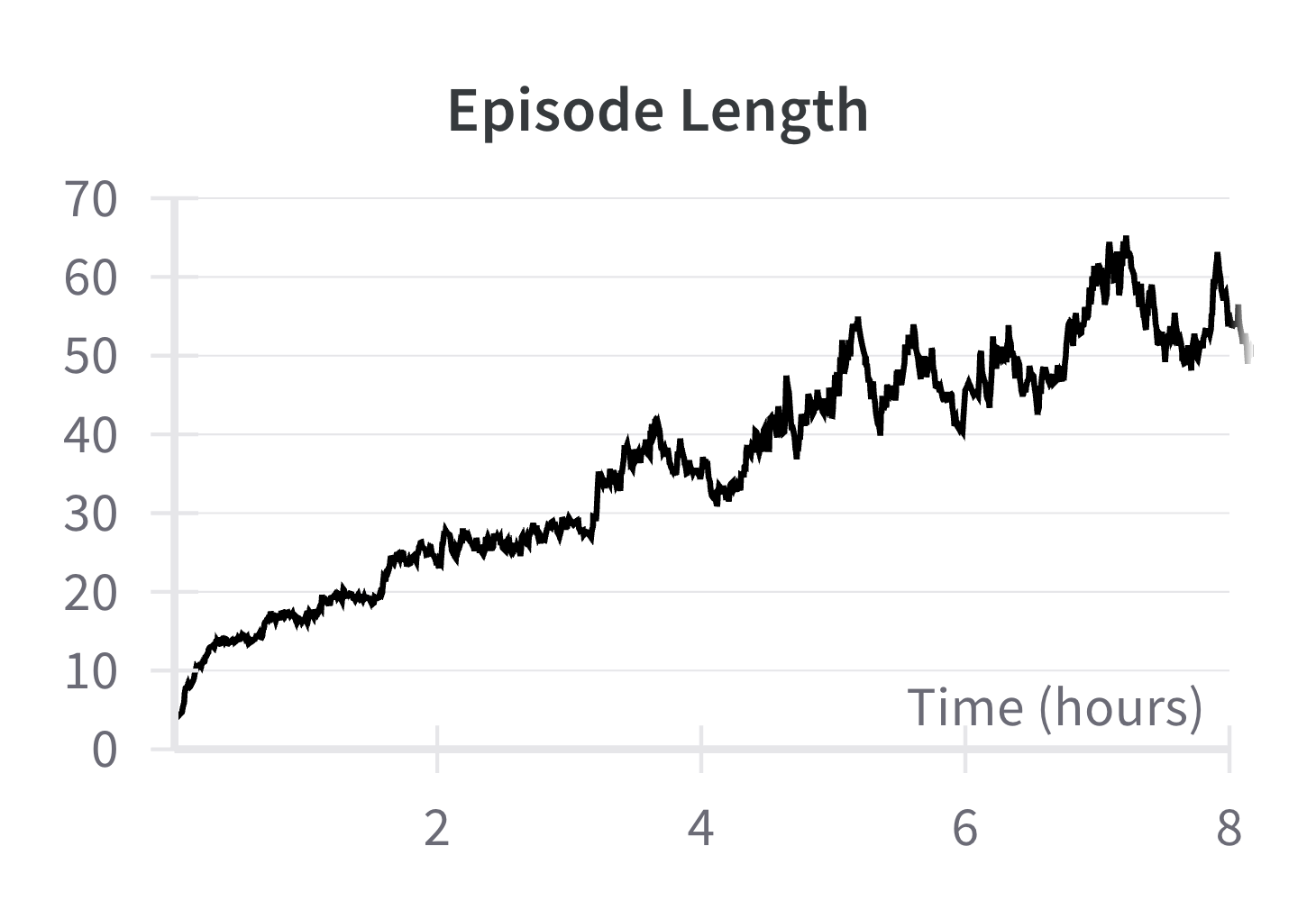}}
    \hfill
    \caption{Visualizations of the evolution of the reward metric averaged over the agents and episode length during training on the advection equation.}
    \label{fig:advection_training_graphs}
\end{figure}

\begin{figure}[h]
    \centering
    \centering
    \hfill
    \raisebox{-\height}{\includegraphics[width=0.4\linewidth]{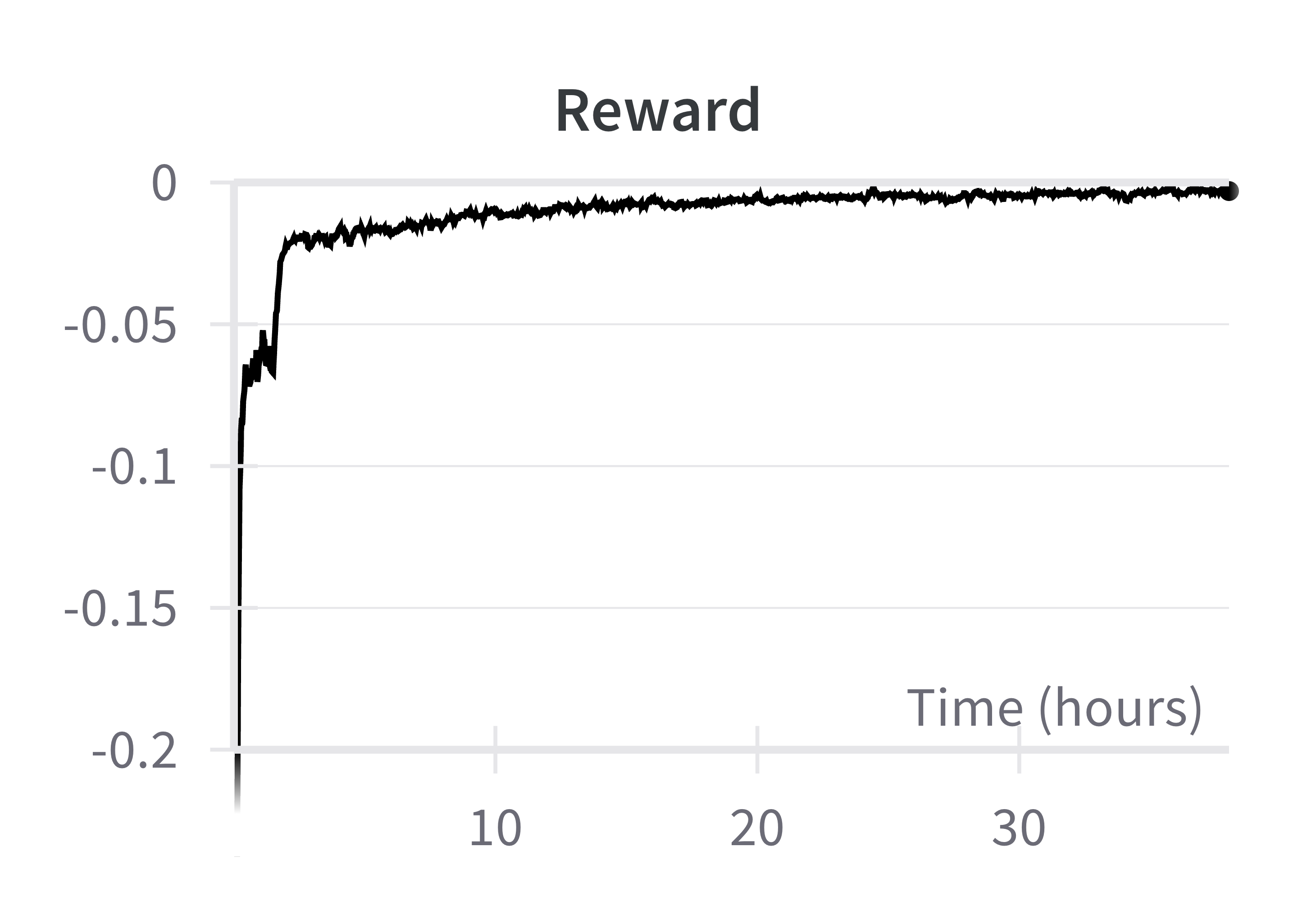}}
    \hfill
    \raisebox{-\height}{\includegraphics[width=0.4\linewidth]{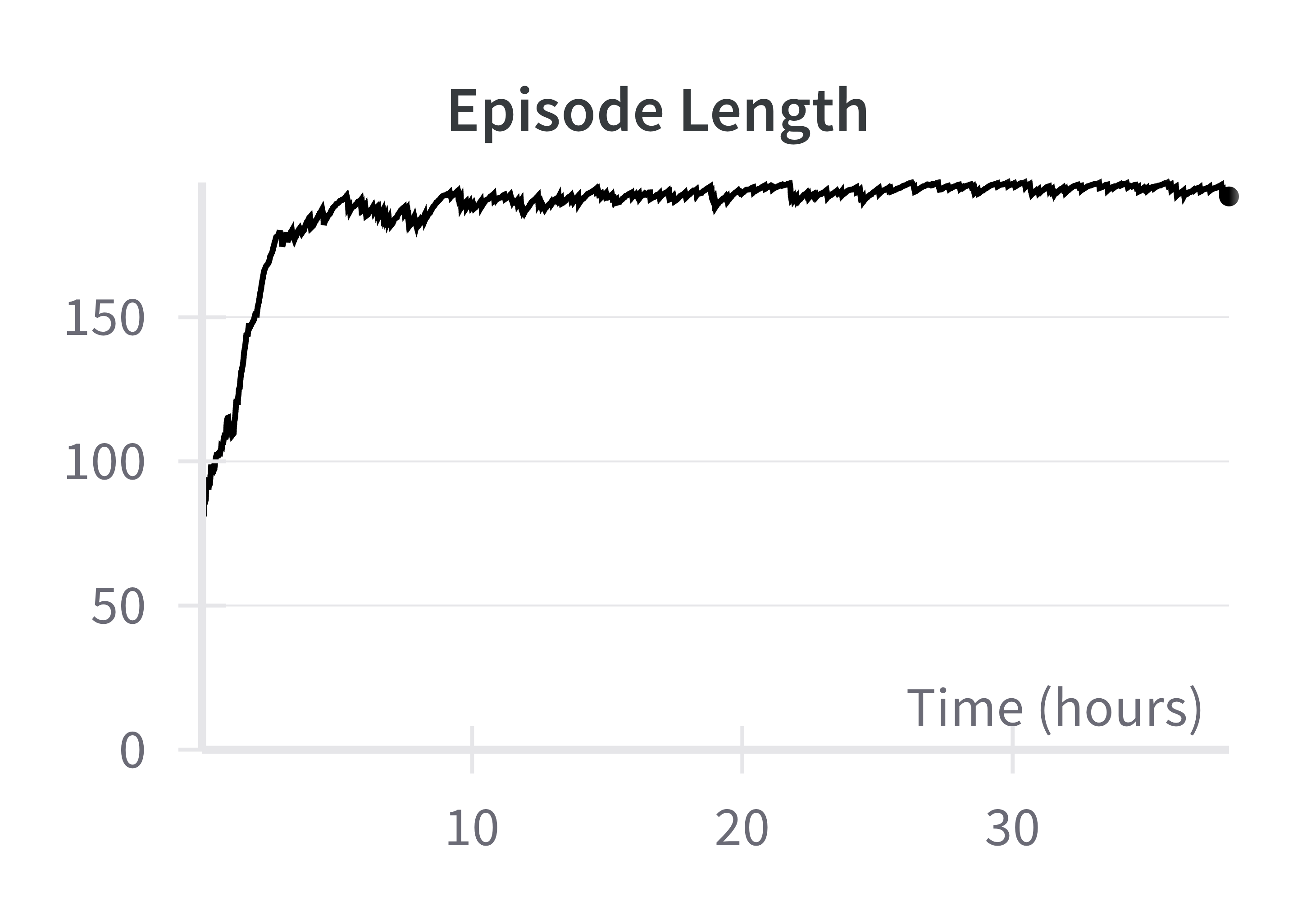}}
    \hfill
    \caption{Visualizations of the evolution of the reward metric averaged over the agents and episode length during training on the Burgers' equation.}
    \label{fig:burgers_training_graphs}
\end{figure}

During training, we use entropy regularization in the PPO objective with a factor of $0.1$ and $0.05$ for the advection and Burgers' equation respectively to encourage exploration. The discount factor is set to $0.95$ and the learning rate to $1\cdot 10^{-5}$. Training is done over 2000 epochs. In each epoch, 1000 transitions are collected. One policy network update is performed after having collected one new episode. We use a batch size of 10 for training. The total number of trainable weights amounts to $188,163$ and the entire training procedure took about 8 hours for the advection equation on an Nvidia A100 GPU. For the Burgers' equation, training took about $30$ hours on the same hardware. We save the policy every 50 epochs and log the corresponding MAE between CGS and FGS after 50 time steps. For evaluation on the advection equation, we chose the policy from epoch 1500 because it had the lowest logged MAE value. \cref{fig:advection_training_graphs} and \cref{fig:burgers_training_graphs} show the reward curves and evolutions of episode lengths. As expected, the episode length increases as the agents become better at keeping the CGS and FGS close to each other. 

\begin{table}[!ht]
\caption{Numerical values used for the advection CGS and FGS. Note that $\widetilde{\Delta t}$ is chosen to guarantee that the CFL condition in the CGS is fulfilled.}
\vskip 0.15in
\centering
\begin{tabular}{@{}l|c|l@{}}
\toprule
$\Omega$ & \multicolumn{2}{c}{$[0, 1] \times [0, 1]$} \\ 
$\tilde N_x$, $\tilde N_y$ & \multicolumn{2}{c}{64}  \\ 
$d, d_t$ &  \multicolumn{2}{c}{4}\\ 
\bottomrule
\multicolumn{3}{c}{Resulting other values:}  \\
\toprule
$ N_x$, $ N_y$ &   $d\cdot\tilde N_x,d\cdot \tilde N_y$ & $=256$ \\ 
$\Delta x, \Delta y$ &  $1/N_{x},1/N_y$ & $\approx 0,0039$  \\ 
$\widetilde{\Delta x}, \widetilde{\Delta y}$ &  $1/{\tilde N_{x}}, 1/{\tilde N_y}$ & $\approx 0,0156$   \\
$\widetilde{\Delta t}$ &  $0.9 \cdot \min(\widetilde{\Delta x}, \widetilde{\Delta y})$& $ \approx 0,0141$  \\
${\Delta t}$ &  $\widetilde{\Delta t}/{d_t}$  & $ \approx 0,0035$  \\ 
\bottomrule
\multicolumn{3}{c}{Discretization schemes:}  \\
\toprule
FGS, Space & \multicolumn{2}{c}{Central difference} \\
FGS, Time & \multicolumn{2}{c}{Fourth-order Runge-Kutta} \\
CGS, Space & \multicolumn{2}{c}{Upwind} \\
CGS, Time & \multicolumn{2}{c}{Forward Euler} \\
\bottomrule
\end{tabular}
\label{tab:advection_params}
\end{table}

\begin{table}[!ht]
\caption{Numerical values used for the Burgers' CGS and FGS. Again, $\widetilde{\Delta t}$ is chosen to guarantee that the CFL condition in the CGS is fulfilled.}
\vskip 0.15in
\centering
\begin{tabular}{@{}l|c|l@{}}
\toprule
$\Omega$ & \multicolumn{2}{c}{$[0, 1] \times [0, 1]$} \\ 
$\tilde N_x$, $\tilde N_y$ & \multicolumn{2}{c}{30}  \\ 
$d, d_t$ &  \multicolumn{2}{c}{5, 10}\\ 
\bottomrule
\multicolumn{3}{c}{Resulting other values:}  \\
\toprule
$ N_x$, $ N_y$ &   $d\cdot\tilde N_x,d\cdot \tilde N_y$ & $=150$ \\ 
$\Delta x, \Delta y$ &  $1/N_{x},1/N_y$ & $\approx 0,0067$  \\ 
$\widetilde{\Delta x}, \widetilde{\Delta y}$ &  $1/{\tilde N_{x}}, 1/{\tilde N_y}$ & $\approx 0,0333$   \\
$\widetilde{\Delta t}$ &  $0.9 \cdot \min(\widetilde{\Delta x}, \widetilde{\Delta y})$ & $ = 0,03$  \\
${\Delta t}$ &  $\widetilde{\Delta t}/{d_t}$ & $ = 0,003$  \\ 
\bottomrule
\multicolumn{3}{c}{Discretization schemes:}  \\
\toprule
FGS, Space & \multicolumn{2}{c}{Upwind} \\
FGS, Time & \multicolumn{2}{c}{Forward Euler} \\
CGS, Space & \multicolumn{2}{c}{Upwind} \\
CGS, Time & \multicolumn{2}{c}{Forward Euler} \\
\bottomrule
\end{tabular}
\label{tab:burgers_params}
\end{table}

\newpage
\subsection{Receptive Field of FCN}
In our Closure-RL problem setting, the receptive field of the FCN corresponds to the observation $\boldsymbol{O}_{ij}$ the agent at point $(i, j)$ is observing. In order to gain insight into this, we analyze the receptive field of our chosen architecture.

In the case of the given IRCNN architecture, the size of the receptive field (RF) of layer $i$ can be recursively calculated given the RF of layer afterward with
\begin{align}
    \text{RF}_{i+1} 
    &= \text{RF}_{i} + (\text{Kernel Size}_{i+1} - 1) \cdot \text{Dilation}_{i+1}\\
    &= \text{RF}_{i} + 2 \cdot \text{Dilation}_{i+1}.\\
\end{align}
The RF field of the first layer $\text{RF}_1$ is equal to its kernel size. By then using the recursive rule, we can calculate the RF at each layer and arrive at a value of $\text{RF}_7 = 33$ for the entire network. From this, we now arrive at the result that agent $(i,j)$ sees a $33 \times 33$ patch of the domain centered around its own location.
\begin{table}[h] 
\caption{Hyperparameters of each of the convolutional layers of the neural network architecture used for the advection equation experiment and the resulting receptive field (RF) at each layer. For the Burgers' equation experiment the architecture is simply adapted by setting the number of in channels of \texttt{Conv2D\_1} to 2 and the number of out channels of $\texttt{Conv2D}\_\pi$ to 4.} 
\vskip 0.15in
\centering
\begin{tabular}{|c|c|c|c|c|c|c|}
\hline
Layer & In Channels & Out Channels & Kernel & Padding & Dilation & RF \\
\hline
\texttt{Conv2D\_1} & 3 & 64 & 3 & 1 & 1 & 3\\
\hline
\texttt{Conv2D\_2} & 64 & 64 & 3 & 2 & 2 & 7 \\
\hline
\texttt{Conv2D\_3} & 64 & 64 & 3 & 3 & 3 & 13 \\
\hline
\texttt{Conv2D\_4} & 64 & 64 & 3 & 4 & 4 & 21 \\
\hline
\texttt{Conv2D\_5} & 64 & 64 & 3 & 3 & 3 & 27 \\
\hline
\texttt{Conv2D\_6} & 64 & 64 & 3 & 2 & 2 & 31 \\
\hline
\texttt{Conv2D}\_$\pi$ & 64 & 2 & 3 & 1 & 1 & 33 \\
\hline
\texttt{Conv2D}\_$\mathcal V$ & 64 & 1 & 3 & 1 & 1 & 33 \\
\hline
\end{tabular}
\label{tab:ircnn}
\end{table}

\subsection{Diverse Velocity Field Generation}
\subsubsection{Distribution for Training}\label{sec:train_vortices}
For the advection equation experiment, the velocity field is randomly generated by taking a linear combination of Taylor-Greene vortices and an additional random translational field. Let $\boldsymbol u_{ij}^{TG,k}, \boldsymbol v_{ij}^{TG,k}$ be the velocity components of the Taylor Greene Vortex with wave number $k$ that are defined as
\begin{align}
    \boldsymbol u_{ij}^{TG,k} &:= \cos(k\boldsymbol x_{i}) \cdot \sin(k \boldsymbol y_j) \\
    \boldsymbol v_{ij}^{TG,k} &:= - \sin(k\boldsymbol x_i) \cdot \cos(k\boldsymbol y_h).
\end{align}
Furthermore, define the velocity components of a translational velocity field as $u^{TL}, v^{TL} \in \mathbb R$. To generate a random incompressible velocity field, we sample 1 to 4 $k$'s from the set $\{1, ..., 6\}$.
For each $k$, we also sample a $\text{sign}_k$ uniformly from the set $\{-1, 1\}$ in order to randomize the vortex directions. For an additional translation term, we sample $u^{TL}, v^{TL}$ independently from  $\text{uniform}(-1, 1)$. We then initialize the velocity field to
\begin{align}
    \boldsymbol u_{ij} &:= u^{TL} + \sum_k \text{sign}_k \cdot \boldsymbol u_{ij}^{TG,k}\\
    \boldsymbol v_{ij} &:= v^{TL} + \sum_k \text{sign}_k \cdot \boldsymbol v_{ij}^{TG,k}.
\end{align}
We will refer to this distribution of vortices as $\mathcal D^{Vortex}_{Train}$. 

For the Burgers' equation experiment, we make some minor modifications to the sampling procedure. The sampling of translational velocity components is omitted and $2$ to $4$ $k$'s are sampled from $\{2, 4, 6, 8\}$. The latter ensures that the periodic boundary conditions are fulfilled during initialization which is important for the stability of the simulations.

\subsubsection{Distribution for Testing} \label{sec:test_vortices}
First, a random sign \( \text{sign} \) is sampled from the set \(\{-1, 1\}\). Subsequently, a scalar \( a\) is randomly sampled from a uniform distribution bounded between 0.5 and 1. The randomization modulates both the magnitude of the velocity components and the direction of the vortex, effectively making the field random yet structured. The functional forms of \(\boldsymbol u_{ij}^{C}\) and \(\boldsymbol v_{ij}^{C}\) are then expressed as
\begin{align}
    \boldsymbol u_{ij}^{V} &:= \text{sign} \cdot  a \cdot \sin^2(\pi \boldsymbol x_i) \sin(2\pi \boldsymbol y_j) \\
    \boldsymbol v_{ij}^{V} &:= - \text{sign} \cdot a \cdot\sin^2(\pi \boldsymbol y_j) \sin(2\pi \boldsymbol x_i).
\end{align}
In the further discussion, we will refer to this distribution of vortices as $\mathcal D^{Vortex}_{Test}$.

\subsection{Computational Complexity}
To quantitatively compare the execution times of the different simulations, we measure the runtime of performing one update step of the environment and report them in \cref{tab:execution_times}. As expected, Closure-RL increases the runtime of the CGS. However, it stays below the FGS times by at least a factor of $5$. The difference is especially pronounced in the example of the advection equation, where the FGS uses a high order scheme on a fine grid, which leads to an execution time difference between Closure-RL and FGS of more than an order of magnitude.
\begin{table}[t] 
\caption{Runtime of one simulation step in ms of the different simulations averaged over 500 steps.}
\label{tab:execution_times}
\vskip 0.15in
\begin{center}
\begin{scriptsize}
\begin{sc}
\begin{tabular}{l|cccr}
\toprule
 & CGS & ACGS & Closure-RL & FGS \\
\midrule
Advection    & $0.31 \pm 0.00$& $ 0.96 \pm 0.00$& $ 2.66 \pm 0.96$ & $ 89.52 \pm 0.47$ \\
Burgers' & $ 0.25 \pm 0.00$ & -& $ 1.82 \pm 0.01$ & $ 10.16 \pm 0.03$\\
\bottomrule
\end{tabular}
\end{sc}
\end{scriptsize}
\end{center}
\vskip -0.1in
\end{table}

We additionally profiled our code for the advection experiment and found that 71.3\% of the runtime is spent on obtaining trajectories from the FGS simulation. Only 1.3 \% of the time is spent on the CGS (excluding the forward pass through the FCN), highlighting the significant computational overhead of the finer grid. The forward passes through the neural network account for 5.0\% of the time, while updating the policy takes another 5.5\%. This is in agreement with the theoretical considerations above that show that the FGS is computationally more expensive than the CGS. We note that there is room for optimization of the numerical solution, as currently the generation of the FGS simulation is not parallelized.

\newpage
\section{Addition of a Global Reward}
\label{sec:glo}
To investigate the effect of adding a global problem specific reward to the reward function, we repeated our experiment for the advection equation and added a penalty based on the deviation from the conservation of the total mass to each local reward term. The deviation from the total mass was scaled so that both parts of the reward term had the same order of magnitude. The obtained mean absolute error reduction compared to CGS was approximately 52 percent after 50 time steps during predictions (the local-only reward resulted in 53 percent error reduction) and thus did not offer any additional benefits compared to our original formulation. The situation may require further investigation for different global constraints and different PDEs. However for the present study the local rewards seem to be sufficient in order to obtain good closures.

\newpage
\section{Discussion of Architecture Choice}
\label{sec:choice}
We have chosen the IRCNN architecture \cite{ircnn} to have a small receptive field such that our agents learn and act locally in a manner that is reminiscent of the numerical discretizations of PDEs. The IRCNN consists of multiple convolutional layers and has been shown to be very successful in image restoration tasks \cite{cv_pixel_rl}. Architectures such as LoFi \cite{khorashadizadehlofi}, which also only have a small receptive field, could be a possible architecture choice for Closure-RL as well, although this would require further investigations.

Other architectures such as U-Nets \cite{ronneberger2015u} or Vision-Transformers \cite{liu2021swin,liang2021swinir,wang2022uformer} are desired to capture multiscale features or long-range dependencies which would significantly increase the receptive field of each agent. Based on our studies regarding the interpretation of actions (see \cref{sec:int}) as well as the derivation of the theoretical optimal action for one of the cases explored in detail (see Section \cref{sec:proof_optimal_action}), these dependencies are generally not needed, as local information should be sufficient to complement numerical schemes. We note that this is not necessarily the case for other systems of interest such as PDEs with forcing terms, PDEs with time-varying parameters or Integro-Differential Equations.

\newpage
\section{Interpretation of Actions}
\label{sec:int}
\begin{figure*}
\centering
    \subfigure{\includegraphics[width=0.65\textwidth]{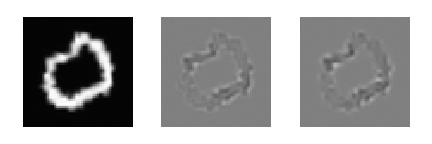}}
    \subfigure{\includegraphics[width=0.24\textwidth]{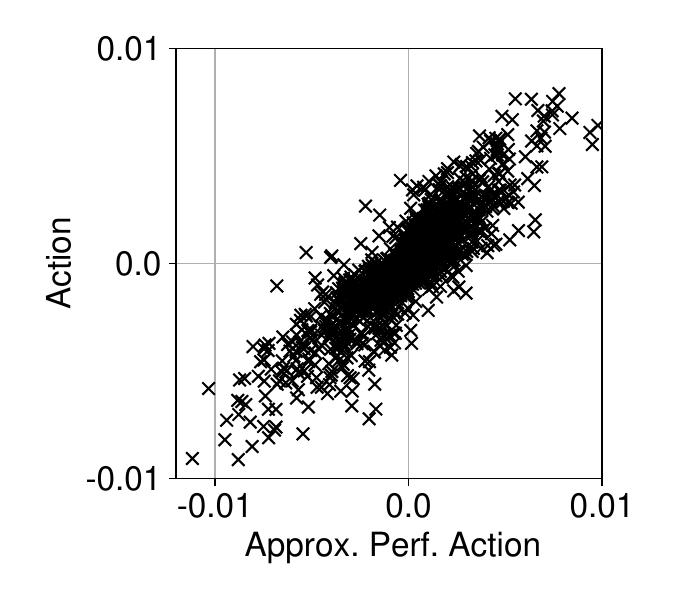}}
    \caption{Visual comparison of the input concentration $\tilde{\boldsymbol \psi}^n$, taken actions $\boldsymbol A^n$ and the corresponding numerical approximation of the optimal action from left to right. The figure on the right-hand side qualitatively shows the linear relation between approximated optimal actions and taken actions.}
    \label{fig:action_interpretation}
\end{figure*}

We are able to derive the optimal update rule for the advection CGS that negates the errors introduced by the numerical scheme (see \cref{sec:proof_optimal_action} for the derivation):
\begin{equation}
\tilde{\boldsymbol\psi}^{n+1} = \mathcal G (\tilde{\boldsymbol\psi}^{n} - \widetilde{\Delta t} \tilde{\boldsymbol\epsilon}^{n-1}, \tilde C^n),
\end{equation}
Here, $ \tilde{\boldsymbol\epsilon}^{n-1} \in \mathbb R^{\tilde N_x \times \tilde N_y}$ is the truncation error of the previous step. However, note that there exists no closed-form solution to calculate the truncation error $\tilde{\boldsymbol\epsilon}^{n-1}$ if only the state of the simulation is given.\\
We therefore employ a numerically approximate for the truncation errors for further analysis:
\begin{equation}
\tilde \epsilon^{n} = \frac{\widetilde{\Delta t}}{2} \tilde \psi_{tt}^n + |\tilde u^n| \frac{\widetilde{\Delta x}}{2}\tilde \psi_{xx}^n + |\tilde v^n| \frac{\widetilde{\Delta y}}{2}\tilde \psi_{yy}^n + \mathcal O(\widetilde{\Delta t}^2, \widetilde{\Delta x}^2, \widetilde{\Delta y}^2)
\end{equation}
Here $\tilde \psi_{xx}^n$,$\tilde \psi_{yy}^n$ and $\tilde \psi_{tt}^n$ are second derivatives of $\psi$ that are numerically estimated using second-order central differences.\\
We compare the obtained optimal update rule for the CGS with the predicted mean action $\boldsymbol \mu_A$ of the policy and thus the learned actions. \cref{fig:action_interpretation} visualizes an example, which indicates that there is a strong linear relationship between the predicted mean action and the respective numerical estimate of the optimal action. \\
In order to further quantify the similarity between the numerical estimates of $- \widetilde{\Delta t} \tilde{\epsilon}^{n-1}_{i,j}$ and the taken action, we compute the Pearson product-moment correlation coefficient for 100 samples. The results are presented in \cref{tab:action_correlation} and show that the learned actions as well as the optimal action of the CGS are highly correlated for all different combinations of seen and unseen ICs and PDEPs. 
\begin{table}[!ht]
\caption{Mean and standard deviation of Pearson product-moment correlation coefficients between $\boldsymbol \mu_A^n$ and numerically approximated $- \widetilde{\Delta t} \tilde{\boldsymbol\epsilon}^{n-1}$ for 100 samples for different combinations of velocity fields and initializations of the concentration.}
\vskip 0.15in
\centering
\begin{tabular}{@{}l|cr@{}}
\toprule
 & MNIST & F-MNIST  \\
\hline 
$\mathcal D^{Vortex}_{Train}$ & $0.82 \pm 0.05$ & $0.70 \pm 0.08$  \\ 
$\mathcal D^{Vortex}_{Test}$ & $0.82 \pm 0.03$ & $0.72 \pm 0.08$ \\ \bottomrule
\end{tabular}
\label{tab:action_correlation}
\end{table}

 Additionally, we note that the truncation error contains a second-order temporal derivative. At first glance, it might seem surprising that the model would be able to predict this temporal derivative as the agents can only observe the current time step. However, the observation contains both the PDE solution, i.e. the concentration for the present examples, as well as the PDEPs, i.e. the velocity fields. Thus, both the concentration and velocity field are passed into the FCN and due to the velocity fields enough information is present to infer this temporal derivative.

%We conclude that for the present example we have shown that the agents are learning actions that are highly corrrelated with what we would expect while still incorporating the property of long-term stable behavior. 
%The result that the high correlation between predicted and appxorimately optimal action persists for out-of-distribution cases, suggests that the model has learned a relatively robust representation of the forcing terms in the CGS. 

%It is furthermore noteworthy that one of the common turbulence models for fluid simulations, the eddy-viscosity model \cite{}, represents the subgrid-scale dynamics with a forcing term that consists of a linear combination of spatial derivatives. All in all, this observed structural similarity is encouraging, as it suggests that Closure-RL is able to learn something of the complexity of the eddy-viscosity model.

\newpage
\section{Proof: Theoretically Optimal Action} \label{sec:proof_optimal_action}
To explore how the actions of the agents can be interpreted, we analyze the optimal actions based on the used numerical schemes. The \textit{perfect} solution would fulfill 
$$
 \mathcal{M}( \tilde{\boldsymbol \psi^n}) +\tilde{ \boldsymbol \epsilon^n}  = 0.
$$

Here, $\mathcal{M}$ is the numerical approximation of the PDE an  $\tilde{\boldsymbol \epsilon^n}$ is the truncation error.

We refer to the numerical approximations of the derivatives in the CGS as $\mathcal{T}^{FE}$ for forward Euler and $\mathcal{D}^{UW}$ for the upwind scheme. We obtain
\begin{align}
    0 &=  \mathcal{M}(\tilde{\boldsymbol\psi^n})+ \tilde{\boldsymbol\epsilon}^n\\
     &=\mathcal{T}^{FE}(\tilde{\boldsymbol\psi}^n, \tilde{\boldsymbol\psi}^{n+1}) + \tilde{\boldsymbol u}^n \mathcal{D}^{UW}_x(\tilde{\boldsymbol\psi}^n) + \tilde{\boldsymbol v}^n \mathcal{D}^{UW}_y(\tilde{\boldsymbol\psi}^n) + \tilde{\boldsymbol\epsilon}^n \\
     &=\frac{\tilde{\boldsymbol\psi}^{n+1} - \tilde{\boldsymbol\psi}^n}{\widetilde{\Delta t}} + \tilde{\boldsymbol u}^n \mathcal{D}^{UW}_x(\tilde{\boldsymbol\psi}^n) + \tilde{\boldsymbol v}^n \mathcal{D}^{UW}_y(\tilde{\boldsymbol\psi}^n) + \tilde{\boldsymbol\epsilon}^n. \\
\end{align}

By rewriting, we obtain the following time stepping rule
\begin{align}
   \tilde{\boldsymbol\psi}^{n+1} 
   &= \tilde{\boldsymbol\psi}^n - \widetilde{\Delta t} \left(\tilde{\boldsymbol u}^n \mathcal{D}^{UW}_x(\tilde{\boldsymbol\psi}^n) + \tilde{\boldsymbol v}^n \mathcal{D}^{UW}_y(\tilde{\boldsymbol\psi}^n) + \tilde{\boldsymbol\epsilon}^n\right)\\
   &= \tilde{\boldsymbol\psi}^n - \widetilde{\Delta t} \left(\tilde{\boldsymbol u}^n \mathcal{D}^{UW}_x(\tilde{\boldsymbol\psi}^n) + \tilde{\boldsymbol v}^n \mathcal{D}^{UW}_y(\tilde{\boldsymbol\psi}^n)\right) - \widetilde{\Delta t} \tilde{\boldsymbol \epsilon}^n\\
   &= \mathcal G(\tilde{\boldsymbol\psi}^n, \tilde C^n) - \widetilde{\Delta t} \tilde{\boldsymbol\epsilon}^n.
\end{align}

where $\tilde C^n:=(\tilde{\boldsymbol u}^n, \tilde{\boldsymbol v}^n)$. This update rule would theoretically find the \textit{exact} solution. It involves a coarse step followed by an additive correction after each step. We can define $ \tilde{\boldsymbol\phi}^{n+1} :=\tilde{\boldsymbol\psi}^{n+1} + \widetilde{\Delta t} \tilde{\boldsymbol\epsilon}^n$, to bring the equation into the same form as seen in the definition of the RL environment
$$ 
\tilde{\boldsymbol\phi}^{n+1} = \mathcal G(\tilde{\boldsymbol\phi}^{n} - \widetilde{\Delta t} \tilde{\boldsymbol\epsilon}^{n-1}, \tilde C^n),
$$
which illustrates that the optimal action at step $n$ would be the previous truncation error times the time increment
\begin{equation}
    \boxed{
    {\boldsymbol A^n}^* = \widetilde{\Delta t} \tilde{\boldsymbol\epsilon}^{n-1}.
    }
\end{equation}

\end{document}